\documentclass[lettersize,journal]{IEEEtran}
\usepackage{amsmath,amsfonts}
\usepackage{algorithmic}
\usepackage{algorithm}
\usepackage{array}
\usepackage[caption=false,font=normalsize,labelfont=sf,textfont=sf]{subfig}
\usepackage{textcomp}
\usepackage{stfloats}
\usepackage{url}
\usepackage{verbatim}
\usepackage{graphicx}
\usepackage{cite}
\usepackage{hyperref}
\usepackage{microtype} 

\usepackage{tabularx,booktabs,multirow,array}
\usepackage{array}                
\usepackage[table]{xcolor}        
\usepackage{booktabs}             

\begin{document}

\title{Stylistic-STORM (ST-STORM) : Perceiving the Semantic Nature of Appearance}

\author{Hamed~Ouattara,
        Pierre~Duthon,
        Pascal~Houssam~Salmane,
        Fr\'ed\'eric~Bernardin,
        and Omar~Ait~Aider%
\thanks{Hamed Ouattara, Pierre Duthon, and Fr\'ed\'eric Bernardin are with the Intelligent Transportation Systems Research Team, Cerema, 8 rue Bernard Palissy, 63017 Clermont-Ferrand, France (e-mail: ouattarahamed6639@gmail.com; pierre.duthon@cerema.fr; frederic.bernardin@cerema.fr).}%
\thanks{Pascal Houssam Salmane is with the Intelligent Transportation Systems Research Team, Cerema, 1 avenue du Colonel Roche, 31400 Toulouse, France (e-mail: pascal.salmane@cerema.fr).}%
\thanks{Omar Ait Aider is with Clermont Auvergne INP, Universit\'e Clermont Auvergne, Institut Pascal, CNRS, 4 Avenue Blaise Pascal, 63178 Clermont-Ferrand, France (e-mail: omar.ait-aider@uca.fr).}%
}

\maketitle

\begin{abstract}

One of the dominant paradigms in self-supervised learning (SSL), illustrated by MoCo or DINO, aims to produce \emph{robust} representations by capturing \emph{features} that are insensitive to certain image transformations such as illumination, reflectance, or geometric changes. This strategy is appropriate when the objective is to recognize objects independently of their appearance (a cat remains a cat despite fur color, lighting, rain, or snow). However, it becomes counterproductive as soon as \emph{appearance} itself constitutes the discriminative signal. In weather analysis, for example, rain streaks, snow granularity, atmospheric scattering, as well as reflections and halos, are not noise: they carry the essential information. In critical applications such as autonomous driving, ignoring these cues is risky, since grip and visibility depend directly on ground conditions and atmospheric conditions. We introduce \textbf{ST-STORM}, a hybrid SSL framework that treats appearance (\emph{style}) as a semantic modality to be \emph{disentangled} from content. Our architecture explicitly separates two latent streams, regulated by \emph{gating} mechanisms. The \emph{Content} branch aims at a stable semantic representation through a JEPA scheme coupled with a contrastive objective, promoting invariance to appearance variations. In parallel, the \emph{Style} branch is constrained to capture appearance signatures (textures, contrasts, scattering) through \emph{feature} prediction (Style-JEPA) and reconstruction under an adversarial constraint. We evaluate \textbf{ST-STORM} on several tasks, including object classification (ImageNet-1K), fine-grained weather characterization, and melanoma detection (ISIC 2024 Challenge). The results show that the \emph{Style} branch effectively isolates complex appearance phenomena (F1=97\% on Multi-Weather and F1=94\% on ISIC 2024 with 10\% labeled data), without degrading the semantic performance (F1=80\% on ImageNet-1K) of the \emph{Content} branch, and improves the preservation of critical appearance information compared with MoCo-v3 and I-JEPA.

\end{abstract}

\begin{IEEEkeywords}
Self-supervised learning, representation disentanglement, style representation, predictive learning, JEPA, appearance modeling, weather analysis, computer vision.
\end{IEEEkeywords}


\section{Introduction}

The search for \emph{universal} visual representations has long been guided by a principle of stability: to be robust, a model must recognize an object regardless of its superficial variations. This idea already appeared in \emph{handcrafted} methods such as the Harris--Stephens repeatable corner detector~\cite{harris1988combined}, SIFT descriptors~\cite{lowe1999object,lowe2004distinctive} (scale and rotation invariance, robustness to moderate photometric variations), and SURF~\cite{bay2006surf}. This philosophy, referred to as \emph{invariance-based learning} by Assran \emph{et al.}~\cite{assran2023self} in their article introducing the I-JEPA model, is now found in modern self-supervised learning (SSL): artificially perturbed views of the same image (croppings, color alterations, geometric transformations) are constrained to project toward a common representation~\cite{simeoni2025dinov3,chen2021empirical}. While this insensitivity often promotes high-level semantic abstraction, it operates by elimination: the network thus learns to treat texture, luminance, and image grain not as descriptive characteristics, but as nuisances to be filtered out in order to reach the object's ``truth.''

\begin{figure}[!t]
    \centering
    \includegraphics[width=0.9\columnwidth]{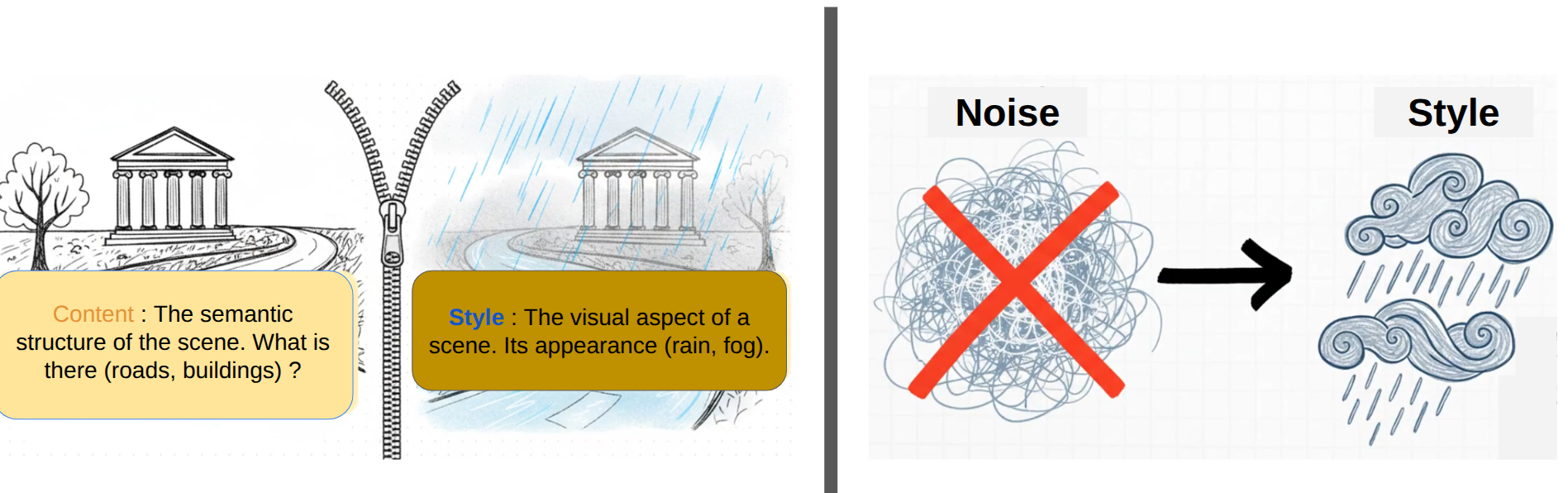}
    \caption{\textbf{Toward a semantic representation of appearance conditions.} The goal is to transform the perception of style: no longer to see it as a random disturbance, but as a structured category. This approach mimics the human ability to dissociate the object (the car) from the context (the rain), while understanding that rain is a physical entity with predictable consequences (slippery road, reduced visibility), which is crucial for decision-making in autonomous driving.}
    \label{fig:Change-noise-semantic}
\end{figure}

However, this bias becomes a critical limitation as soon as one moves beyond ``coarse'' object recognition toward \emph{fine-grained} tasks, where the discriminative information lies precisely in \emph{style} or \emph{texture}. Distinguishing cat breeds (Siamese, Ragdoll, Bombay), for instance, relies on appearance cues (coat patterns, colorimetry, micro-textures), whereas a strongly \emph{invariant} SSL model has learned to attenuate these variations in favor of more generic signals (silhouette, presence of ears, tail). Similarly, in medical imaging, many diagnostic decisions depend on micro-structures and textural signatures (granularity, glandular organization, nuclear irregularities, stromal patterns); in this context, learning to ``ignore'' grain, local contrasts, or certain photometric variations may suppress essential markers for cancer detection or grading. More generally, several works~\cite{xiao2020should,chavhan2023amortised,wang2024understandingroleequivarianceselfsupervised} highlight that the invariance useful for semantic classification (e.g., ``this is a road'') is not necessarily the one required for fine scene analysis. This mismatch is particularly marked in weather analysis: rain streaks, fog diffusion, snow granularity, halos, and reflections are not noise to be removed, but the very signal one seeks to measure (\hyperref[fig:Change-noise-semantic]{Fig.~\ref{fig:Change-noise-semantic}}). It therefore becomes necessary to preserve these appearance cues rather than dilute them into a systematically invariant abstraction.

The partly arbitrary nature of the choice of invariances, and the biases it induces \emph{by construction}, calls into question the ability of such SSL methods to produce truly universal representations. This observation has motivated the emergence of alternative approaches, including \emph{I-JEPA}~\cite{assran2023self}. This so-called \emph{predictive} method differs from contrastive or distillation-based frameworks that explicitly align augmented views: I-JEPA learns by \emph{predicting} the latent representations of masked regions from their context. Although this objective does not directly impose invariance to augmentations, it also provides no explicit mechanism to \emph{preserve} appearance. In practice, prediction naturally favors the most \emph{stable} and \emph{predictable} factors of variation. This bias is reinforced when supervision is defined in the latent space of a deep target encoder with a large receptive field, which tends to \emph{smooth} photometric details: local high-frequency variations may fluctuate strongly without being necessary for global semantic coherence. Thus, in domains where discriminative information lies in fine signals (rain streaks, specularities on wet roads, atmospheric scattering), this implicit preference for ``smoothed'' representations may lead to underestimating appearance components that are nonetheless crucial.

It is precisely this absence of \emph{explicit} consideration of appearance that \emph{ST-STORM} seeks to address. Unlike invariance-based SSL methods, we do not consider style as a nuisance to be removed (\hyperref[fig:Change-noise-semantic]{Fig.~\ref{fig:Change-noise-semantic}}); and unlike predictive frameworks of the I-JEPA type, we introduce dedicated mechanisms to \emph{preserve} and \emph{organize} appearance cues. Our central hypothesis is that appearance (or \emph{style}) \emph{features} constitute a semantic modality in their own right, complementary to content. ST-STORM is therefore \emph{two-headed}: one branch learns a robust content representation that is transferable across styles, while a second branch learns a semantic representation of \emph{appearance}.

\begin{figure}[!t]
    \centering
    \includegraphics[width=0.9\columnwidth]{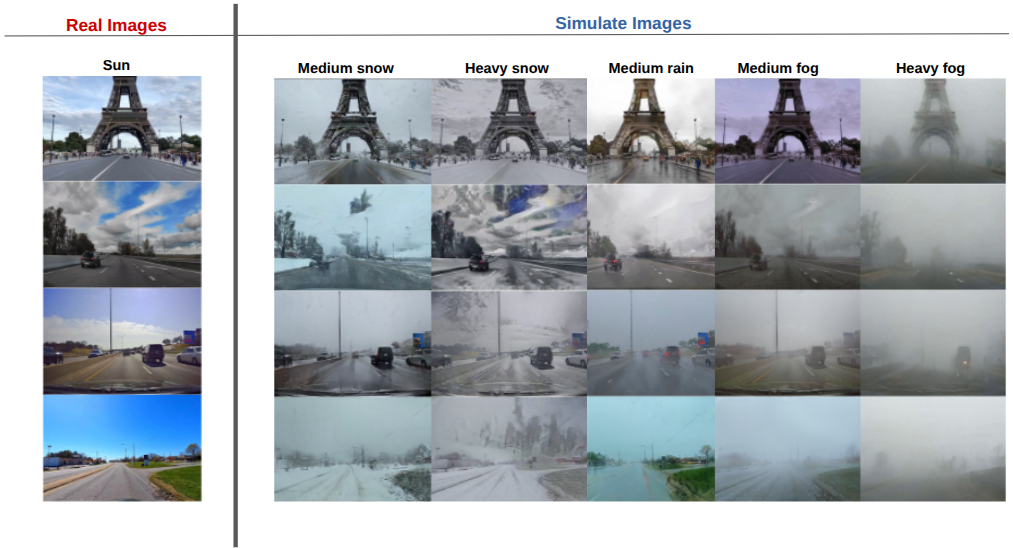}
    \caption{Weather simulation by CycleGAN style transfer~\cite{key2}. Left column: real images (source). Right column: stylized outputs reproducing rain, snow, or fog at different intensity levels. This example shows that an unpaired model can synthesize characteristic weather cues (rain streaks, luminous veil, color drift), which motivates the idea of treating weather as a \emph{style} variation.}
    \label{fig:example-style-transfer}
\end{figure}

This hypothesis is supported by recent progress in generative AI, and in particular in style transfer, which shows that it is possible to strongly alter the appearance of a scene while preserving its structure (content), as illustrated in \hyperref[fig:example-style-transfer]{Fig.~\ref{fig:example-style-transfer}}. Unpaired methods such as CycleGAN~\cite{key2} or CUT~\cite{key3} simulate complex weather conditions (rain, snow, fog) by mainly modifying texture, colorimetry, and diffusion, while preserving the scene ``content.''

We leverage style transfer and spectral perturbations (using the Fast Fourier Transform (FFT)~\cite{cooley1965algorithm} and the Sliced Wasserstein Distance (SWD)~\cite{bonneel2015sliced}) to introduce \emph{stylistic chaos} serving as directed augmentation: from a source image, we synthesize views that preserve \emph{content} while altering \emph{appearance} in a controlled manner. Unlike MAE-style masking--reconstruction approaches~\cite{he2022masked}, which indiscriminately occlude both content and style and aim at pixel reconstruction, our strategy explicitly perturbs photometric and spectral attributes (texture, colorimetry, diffusion) over the whole image in order to promote targeted disentanglement. Appearance signatures are encoded by a \emph{StylePyramidEncoder} that produces multi-scale \emph{tokens} through adaptive average pooling. To prevent reconstructive supervision from encouraging the learning of \emph{contingent} and poorly transferable details, we introduce \emph{Style-JEPA}~\cite{assran2023self}: the model must predict masked style tokens in latent space, which filters out noisy components in favor of stable and semantically relevant appearance factors. In parallel, a \emph{Content-JEPA} branch coupled with a contrastive loss enforces invariance of content \emph{features} to perturbations induced by stylistic chaos. Ultimately, ST-STORM learns two complementary latent spaces: a robust \emph{Content} stream and a semantic \emph{Style} stream dedicated to appearance. Our central contribution concerns this \emph{Style} branch, which we show is able to capture fine and discriminative spectral signatures without degrading the semantic performance of the content branch.

To validate the relevance of our approach, we adopt a standard evaluation protocol based on representation transfer: self-supervised pretraining, freezing of the backbone weights, and then supervised \emph{fine-tuning} using increasing fractions of labeled data (from 1\% to 10\%). We compare \textbf{ST-STORM} to MoCo-v3~\cite{chen2021empirical} and I-JEPA~\cite{assran2023self} on several tasks covering (i) a scenario dominated by \emph{semantic content} and (ii) scenarios dominated by \emph{visual appearance}. For the content-centered task, we evaluate the \emph{Content} branch on an ImageNet-1K classification task, to verify that it remains competitive with the state of the art when the objective is object recognition independently of appearance. For tasks where the discriminative information is mainly stylistic, we evaluate the \emph{Style} branch on fine-grained characterization problems, notably weather attribute classification\footnote{\href{https://github.com/Hamedkiri/Weather\_MultiTask\_Datasets}{https://github.com/Hamedkiri/Weather\_MultiTask\_Datasets}} and melanoma detection\footnote{\href{https://www.kaggle.com/competitions/isic-2024-challenge/data}{https://www.kaggle.com/competitions/isic-2024-challenge/data}}, where the relevant cues rely primarily on texture, color, and visual structure. In particular, for fine-grained weather feature classification, the results show that the \emph{Style} branch (F1 = 97\%) outperforms not only the \emph{Content} branch (F1 = 90\%), but also the reference representations learned by I-JEPA~\cite{assran2023self} (F1 = 87\%) and MoCo-v3~\cite{chen2021empirical} (F1 = 91\%). Finally, we measure the model's generalization ability on the following datasets: \emph{Oxford Flowers-102}, \emph{Oxford-IIIT-Pets}, and \emph{CIFAR100}. The source code of the ST-STORM model is available here: \url{https://github.com/Hamedkiri/RT-STORM-V2}
\section{Related Work}
\label{related-work}

The pursuit of \emph{universal visual representations} is inspired by a simple cognitive intuition: without explicit labels, humans learn to group diverse instances under stable concepts (``cat,'' ``chair,'' ``car'') by capturing what is \emph{common} and relegating accidental variations (lighting, pose, viewpoint) to the background. In other words, they build categories by exploiting the world's regularities, not annotations.

Unsupervised and self-supervised learning in vision pursues the same objective: learning transferable \emph{features} from signals intrinsic to the data. Once pre-trained, an encoder should produce representations that allow efficient adaptation to many downstream tasks with few labeled examples. The difference is that, for machines, these ``regularities'' must be \emph{operationalized} as training objectives: making two views consistent, reconstructing a missing part, or predicting a latent representation. These choices implicitly define which information is considered stable and relevant, or instead contingent.

In this review, we organize the literature around four families of self-supervised objectives corresponding to these principles: (i) \emph{invariance}-based methods, which explicitly enforce insensitivity to certain transformations to promote semantic abstraction; (ii) \emph{reconstruction}-based methods (e.g., masking--reconstruction), which learn representations by reconstructing missing regions; (iii) \emph{predictive} methods, which learn to anticipate latent representations and tend to filter contingent information because it is hard to predict; and (iv) works aiming to learn a \emph{semantic} representation of appearance (\emph{style}) when it itself constitutes transferable information.

\subsection{Invariance-Based Approaches and Their Limitations}
\label{subsec:Invariance-approach}

\begin{figure}[!t]
    \centering
    \includegraphics[width=0.9\columnwidth]{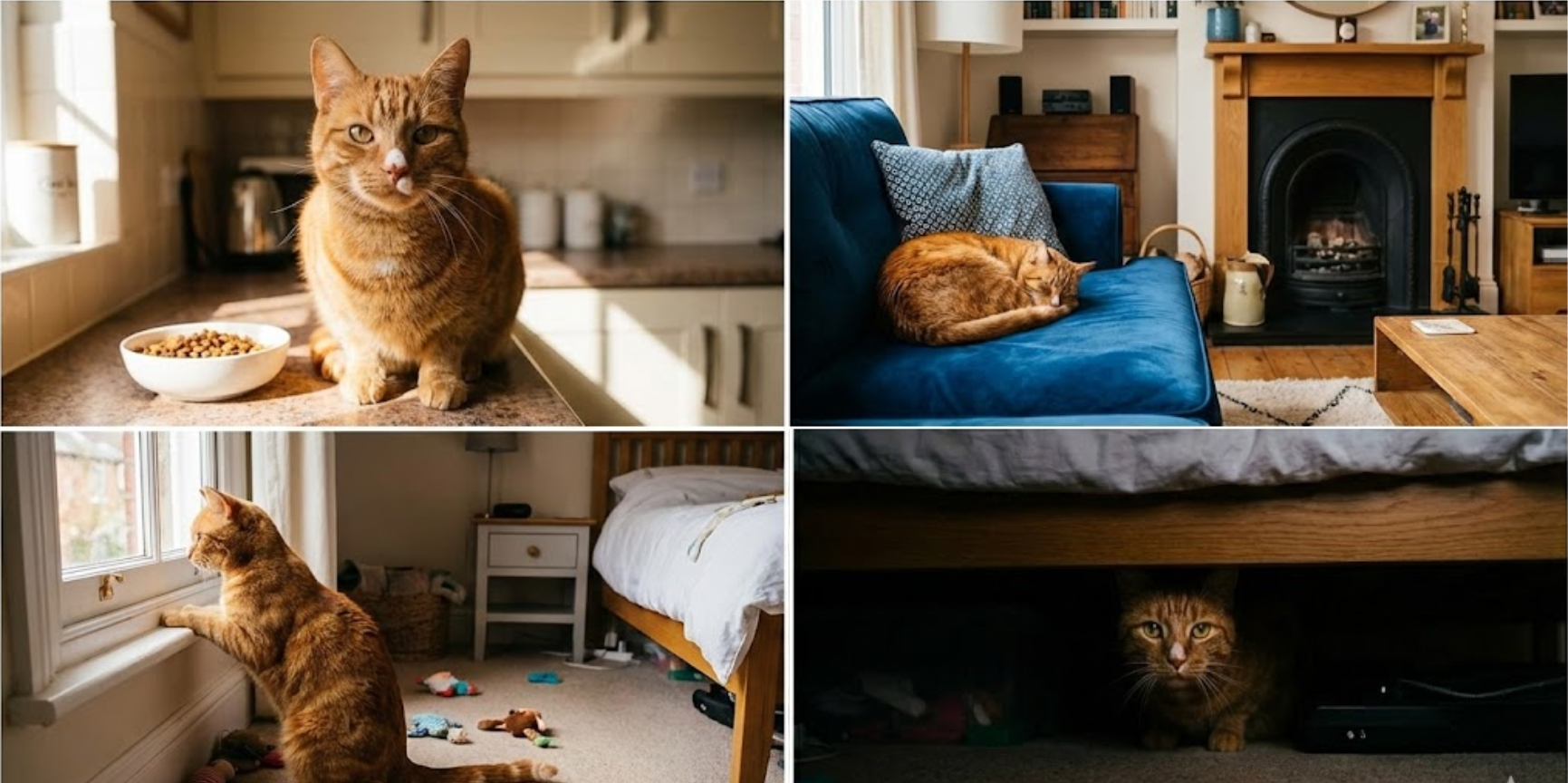}
    \caption{\textit{Illustration of the invariance paradigm in self-supervised learning.} Consider the same cat photographed in an apartment: it moves from the living room to the bedroom, the lighting changes (shadow, lamp, daylight), it adopts different poses, and it may be partially occluded (for example under a bed). Despite these appearance variations, the identity of the object does not change: it is still the same cat. The principle of SSL methods \emph{by invariance} is precisely to force the network to produce \emph{similar} representations for all these views, in order to learn robust and stable \emph{features}. In practice, one typically starts from a single image and generates two views by applying augmentations (cropping, rotation, blur, color jitter, changes in brightness/contrast, etc.). The model is then trained to pull the embeddings of these two views closer together: it thus learns to ignore variations in illumination, color, pose, or occlusion, and to retain what is invariant and semantically relevant.}
    \label{fig:Cats-in-differents-situation}
\end{figure}

The foundational intuition behind SSL \emph{by invariance} is the following: the same entity should keep the same representation despite observation variations (\hyperref[fig:Cats-in-differents-situation]{Fig.~\ref{fig:Cats-in-differents-situation}}). For example, if we photograph \emph{the same cat} outdoors in bright sunlight, then indoors under a chair, then on a bookshelf, the image changes significantly (lighting, pose, viewpoint, partial occlusions), but the semantic identity remains the same. A good encoder should therefore extract \emph{stable} features that recognize it is still the same cat, regardless of these transformations.

Contrastive methods such as MoCo~\cite{he2020momentum,chen2021empirical} and SimCLR~\cite{chen2020simclr}, as well as distillation methods like DINO~\cite{simeoni2025dinov3}, operationalize this idea by generating augmented views of the same image (cropping, color jitter, blur, etc.) and forcing the encoder to produce close representations. Augmentations thus act as ``real-world transformations'' that the model should ignore. This strategy yields highly transferable semantic \emph{features}, explaining the empirical success of MoCo and DINO.

However, this robustness depends on an \emph{a priori} choice of invariances. When the suppressed variations contain the discriminative signal (orientation, color, texture, micro-structures), invariance becomes counter-productive. Several works show that excessive invariance can degrade downstream tasks that instead require \emph{sensitivity} to variations~\cite{xiao2020should,chavhan2023amortised,wang2024understandingroleequivarianceselfsupervised}. Xiao \emph{et al.}~\cite{xiao2020should}, for instance, propose \emph{Leave-one-out Contrastive Learning} (LooC) to preserve certain information in dedicated subspaces. Nevertheless, these approaches generally remain centered on handling \emph{manual} augmentations (often geometric/photometric) and do not explicitly address the case where appearance itself is a semantic modality to preserve.

\subsection{Reconstruction-Based Approaches and the Trap of Contingency}
\label{subsec:Reconstruction-approach}

\begin{figure}[!t]
    \centering
    \includegraphics[width=0.9\columnwidth]{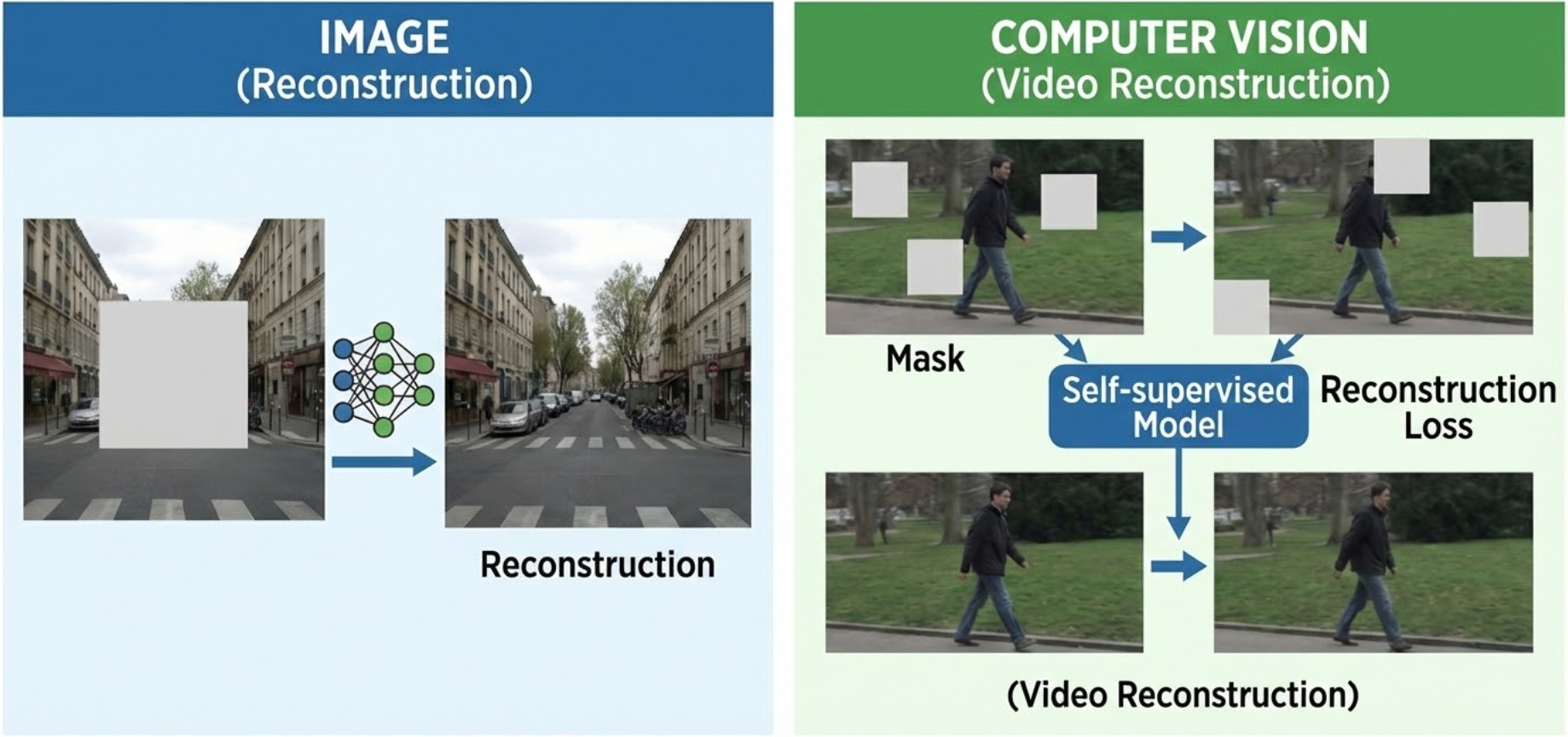}
    \caption{\textit{Principle of reconstruction-based self-supervised approaches.} Part of the image is masked and the network must reconstruct it from the visible context, in order to learn the relations between regions. Limitation: reconstruction may encourage memorization of contingent details (textures/noise), which can be poorly transferable to downstream tasks.}
    \label{fig:Reconstruction-example}
\end{figure}

To escape the biases induced by \emph{a priori} invariance choices, a powerful alternative lies in generative or reconstruction-based approaches, popularized by \emph{Masked Image Modeling} (MIM)~\cite{bao2021beit,he2016deep,he2022masked}. The principle is to learn representations by forcing the model to predict masked parts of the signal (pixels or visual tokens) from their visible context(\hyperref[fig:Reconstruction-example]{Fig.~\ref{fig:Reconstruction-example}}). By solving this dense task, the network implicitly captures geometric structure and spatial dependencies without requiring complex manual augmentations.

However, these methods suffer from a major intrinsic limitation: \emph{over-specification} toward contingent information. As highlighted by recent works~\cite{yang2025pursuit,vanassel2025jointembeddingvsreconstruction,liu2023pixmim,xue2023stare}, constraining a model to reconstruct the raw signal forces it to waste capacity modeling unpredictable and semantically irrelevant details. Consider a face: while context may predict the presence of an eye (semantic information), it is often insufficient to infer the exact iris color or the precise light reflection (contingent information). Penalizing the model for these inevitable reconstruction errors encourages encoding high-frequency details at the expense of semantic abstraction, harming feature transferability.

This is where \emph{ST-STORM} differs. Although our architecture seeks to model appearance (style), we reject pixel-by-pixel reconstruction in favor of a predictive approach in latent space, inspired by I-JEPA. Instead of generating raw textures, our \emph{Style} branch predicts abstract representations of appearance. This predictability constraint acts as a filter: it captures coherent, repeatable style structures (granularity, global spectral distribution) while ignoring stochastic and contingent variations that hamper classical reconstructive approaches.

\subsection{New Perspectives: Predictability over Reconstruction}
\label{subsec:predictictibilite}

To escape the contingency trap inherent to pixel-level reconstruction, a new family of \emph{predictive} methods~\cite{assran2023self,fei2023jepa,assran2025v} has emerged, notably embodied by I-JEPA~\cite{assran2023self}. These approaches rely on a strong philosophical postulate: a concept is defined by what is \emph{predictable} in the signal, not by the entirety of the signal.

\begin{figure}[!t]
    \centering
    \includegraphics[width=0.9\columnwidth]{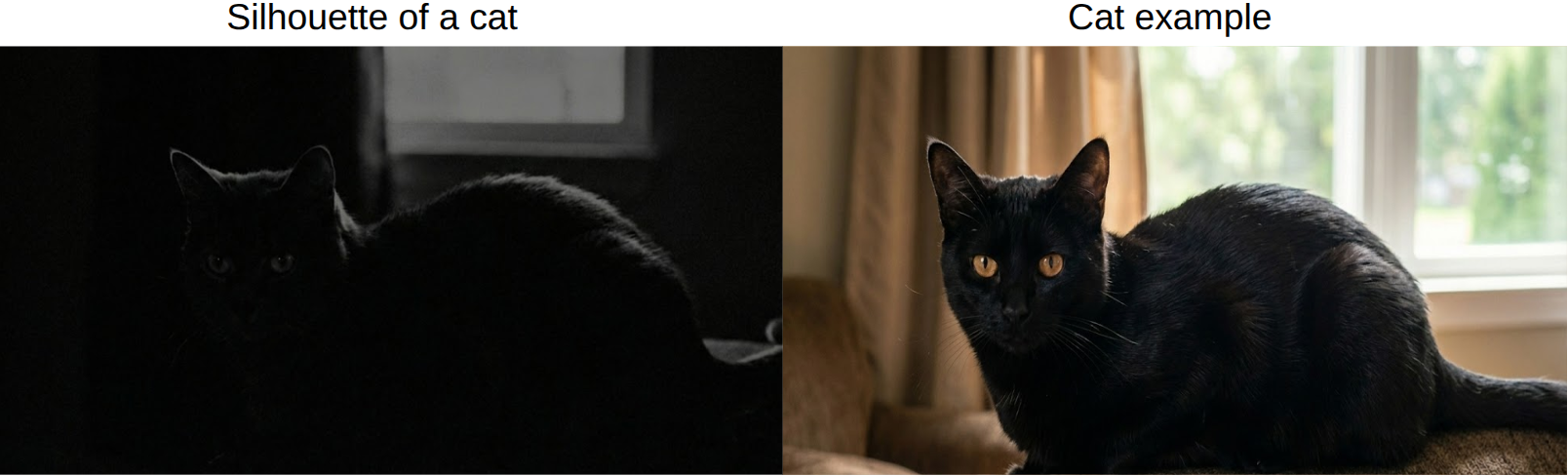}
    \caption{\textit{Principle of predictability at the embedding level (I-JEPA).} From a partial context, some properties of a cat are predictable (presence of ears, whiskers, tail), whereas contingent details are much less so (exact fur color, eye shade). I-JEPA therefore imposes a prediction constraint in the representation space rather than at the pixel level, in order to favor abstract and transferable information. In our setting, if the context provides information about the \emph{breed} (e.g., Bombay), then appearance attributes also become predictable (black coat, copper/golden eyes), illustrating how stylistic cues can acquire semantic significance.}
    \label{fig:Jepa-cat}
\end{figure}

Consider the concept ``cat.'' Our mental representation integrates stable and predictable elements (four legs, whiskers, ear shape). By contrast, given only a partial cat silhouette, it is impossible to guess the exact fur color or the position of every hair ((\hyperref[fig:Jepa-cat]{Fig.~\ref{fig:Jepa-cat}})). These details are contingent information. Where reconstructive methods (MAE) force the network to encode these superfluous details to succeed, predictive methods (JEPA) move the objective into an abstract space. Concretely, from a context block $x$, the model does not seek to hallucinate missing pixels $y$, but to predict their latent representation $s_y$ via a predictor $g_\phi$. Training minimizes a distance in this semantic space:
\[
\mathcal{L} = \left\| g_\phi(f_\theta(x), m) - s_y \right\|_2^2,
\]
mechanically filtering unpredictable noise.

\paragraph{The style paradox: noise or semantic concept?}
The distinction between \emph{contingent} and \emph{predictable} information raises a key question for \emph{ST-STORM}: can we learn a \emph{semantic} representation of appearance (\emph{style})? At first glance, appearance seems dominated by random details (grain, local variations, exact drop location). Yet these details become \emph{predictable} once conditioned by a sub-category or phenomenon. For example, once the cat's breed (e.g., "Bombay") was identified, its black fur and copper eyes ceased to be accidental and became distinctive attributes.

The same reasoning applies to weather: while the exact location of rain streaks is stochastic, the global phenomenon ``rain'' induces stable regularities (scattering, ground reflections, frequency signatures, gradient distributions). Pioneering works have shown that such low-level statistics can suffice for weather inference~\cite{ship2024real,zhao2011feature}. Thus, just as certain visual motifs allow inferring ``cat,'' appearance regularities allow inferring ``rain.'' This is precisely the goal of our \emph{Style-JEPA} branch: not to reconstruct pixels, but to predict a coherent \emph{latent signature} of environmental conditions. Unlike I-JEPA~\cite{assran2023self}, which provides no dedicated mechanism to capture style (and may attenuate these cues through deep encoders, as we will show in Sec.~\ref{sec:background}), ST-STORM is designed to explicitly preserve and structure these stylistic \emph{features} to produce an exploitable semantic representation.

\subsection{Semantic Representations of Style: State of the Art and Limits}
\label{subsec:style-semantic-repr}

Several works have sought to learn reusable \emph{style} representations for downstream tasks (e.g., artistic style, textures, appearance attributes)~\cite{cui2025community,ruta2023aladin,gairola2020unsupervised,zhang2017deep,lin2017bilinearcnnsfinegrainedvisual}. The common goal is to capture \emph{discriminative} appearance signatures while making them as independent as possible from semantic content.

Among recent approaches, Ruta \emph{et al.}~\cite{ruta2023aladin} propose \emph{ALADIN-NST}, which learns an artistic style descriptor by generating on-the-fly a large synthetic dataset via \emph{Neural Style Transfer} (NST). The model is then trained with a contrastive loss: two images rendered with the \emph{same} style form a positive pair, while different styles form negatives. The final representation combines global statistics (from a VGG) and more localized cues (via a ViT).

These methods confirm it is possible to learn a highly informative latent space for style, but they have two important limitations for our use case. First, NST-based pipelines (in the sense of Gatys \emph{et al.}~\cite{gatys2016image}) rely on pre-trained descriptors (often VGG) and on a style/content separation \emph{implicitly} defined by that backbone, which can be fragile outside the artistic domain and limit generalization to real images (more details in Sec.~\ref{sec:background}). Second, these approaches do not necessarily provide an explicit mechanism to filter \emph{contingent} appearance information (random details) in favor of \emph{stable} and predictable factors.

Our approach aligns with this line of work but differs in two ways: (i) we use style transformations closer to real data (e.g., unpaired domain translation such as CycleGAN/CUT~\cite{key2,key3}) to perturb appearance in a controlled manner, and (ii) we impose a \emph{predictability} constraint (\emph{Style-JEPA}) on multi-scale tokens to obtain a \emph{semantic} and transferable style representation, rather than a descriptor merely correlated with reconstruction artifacts.

\section{Background}
\label{sec:background}

\begin{figure}[!t]
  \centering
  \includegraphics[width=\columnwidth]{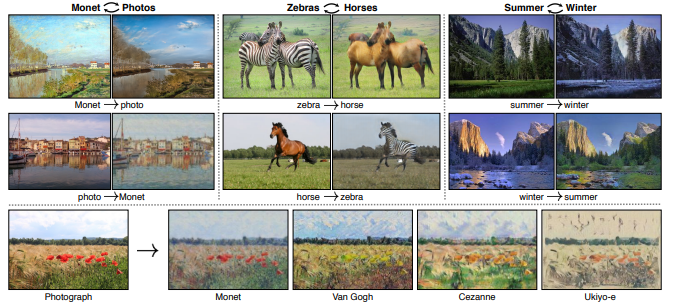}
  \caption{Examples of domain translation by CycleGAN~\cite{key2}, illustrating different forms of appearance transfer.}
  \label{fig:style-transfer-possibilty}
\end{figure}

In computer vision, \emph{style transfer} refers to methods that modify the appearance of an image while preserving,
as much as possible, its structure: the scene and the objects remain recognizable, but their visual rendering changes.
One may thus transform a photograph into a painter’s ``canvas,'' simulate a weather condition,
or convert a horse into a zebra, as illustrated in Fig.~\ref{fig:style-transfer-possibilty}.
In all cases, the problem amounts to deciding \emph{what should remain} and \emph{what may change}.

In this article, we call \emph{content} the information that we seek to preserve, and \emph{style} the information that we allow to vary.
For our purpose, content mainly corresponds to the geometric and semantic organization of the scene
(positions and shapes of roads, vehicles, pedestrians, buildings, markings).
Style gathers the appearance factors that are superimposed on this structure: texture, contrast, colorimetry, illumination,
atmospheric diffusion, reflections, halos, attenuation of details, etc.

This separation is nevertheless not absolute. Appearance conditions what is observable, and may therefore modify what we perceive as content.
For example, by progressively immersing an object in darkness, one changes its appearance, but one also makes certain parts unobservable.
Conversely, changing the observation modality (e.g., RGB $\rightarrow$ infrared) sometimes reveals structures that are invisible in visible light.
In other words, the boundary between style and content depends on the task, the sensor, and the physical acquisition conditions; it cannot be universal.

This question --- where to place the boundary between what should be preserved and what may be transformed --- structures the style transfer literature.
In the following, we distinguish (i) approaches that impose a relatively \emph{fixed} separation between content and style,
and (ii) approaches that learn or control this separation in an \emph{adaptive} manner.
As we will see, this second family is better suited to natural images and to our weather simulation objectives.

\subsection{Fixed-boundary approaches: Neural Style Transfer (NST)}
\label{sub-sec:NST}

\begin{figure}[!t]
  \centering
  \includegraphics[width=0.9\columnwidth]{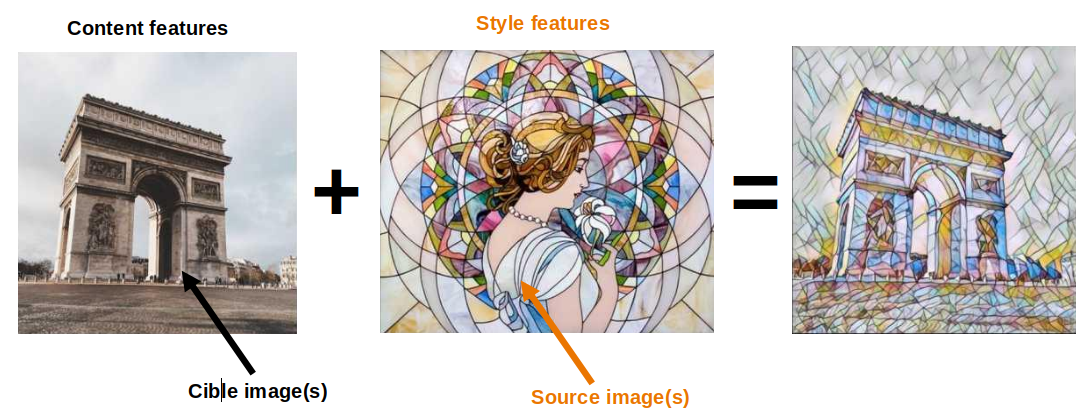}
  \caption{Neural style transfer~\cite{gatys2016image}: a synthetic image is optimized to preserve the \emph{content} of a source image and reproduce the \emph{style} statistics (Gram matrices) of a reference image.}
  \label{fig:figure4}
\end{figure}

\emph{Neural Style Transfer} (NST), popularized by Gatys \emph{et al.}~\cite{gatys2016image} (Fig.~\ref{fig:figure4}), formulates style transfer as the combination of two signals extracted by a recognition network pre-trained in a supervised manner (typically VGG~\cite{simonyan2015deepconvolutionalnetworkslargescale}). The central hypothesis is that the deep layers of such a network mainly encode \emph{content} (semantic structure): they group together visually diverse occurrences of the same category, which makes their representations relatively robust to appearance variations. By contrast, \emph{style} is modeled as a global appearance statistic, captured by the correlations between feature channels within a layer.

\begin{figure}[!t]
  \centering
  \includegraphics[width=\columnwidth]{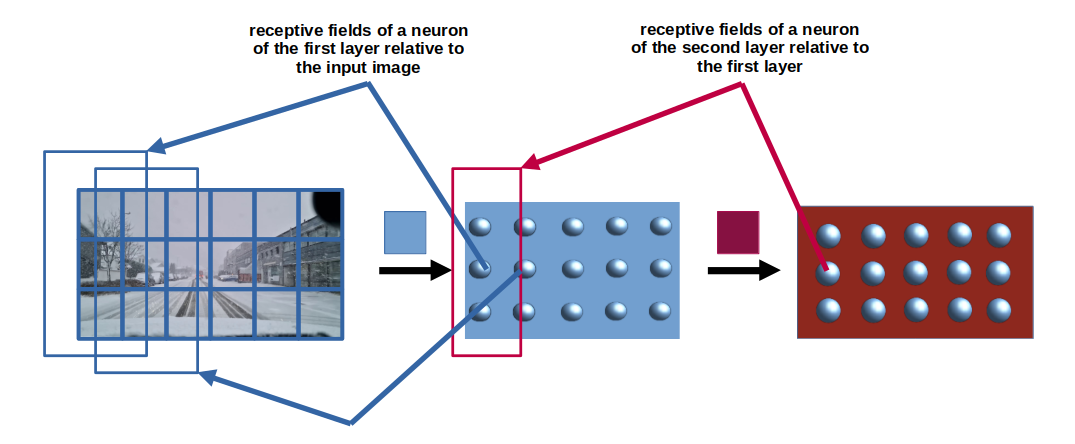}
  \caption{The receptive field of a neuron is the part of the image to which it is directly or indirectly connected. In a convolutional network, the deeper the layer, the broader the spatial aggregation, and thus the larger the receptive field, and the more the descriptors reflect global properties.}
  \label{fig:figure5}
\end{figure}

Concretely, if $F\in\mathbb{R}^{C\times H'\times W'}$ denotes a feature tensor and $X\in\mathbb{R}^{C\times S}$ its flattened version ($S=H'W'$), the Gram matrix
\begin{equation}
G \;=\; \tfrac{1}{S}\, X X^\top \in \mathbb{R}^{C\times C}
\end{equation}
summarizes co-activation statistics that are largely independent of the explicit geometry of objects. In practice, the choice of layers controls the scale of the stylized attributes: intermediate layers describe more local patterns, whereas deep layers reflect more global properties, in line with the increase of the receptive field (Fig.~\ref{fig:figure5}).

This operational definition, which provides a fixed boundary between style and content, nevertheless has two important limitations outside the artistic setting. First, the classical Gram representation loses spatial arrangement (invariance to the permutation of positions), which penalizes the restitution of localized effects (e.g., non-uniform fog). Second, \emph{content} is defined through a supervised backbone: it depends on the biases, categories, and performance of the pre-trained model, and corresponds more to a \emph{semantic} structure than to a \emph{geometric} one. This may lead to preserving the ``class'' while altering the instance (e.g., substituting one road sign for another), which is acceptable in art but problematic in critical contexts such as autonomous driving.

\subsection{Approaches with a learned or controlled boundary: tension between style and content}
\label{subsec:tension-style-content}

Recent \emph{paired} image translation methods (Pix2Pix~\cite{key1}) and \emph{unpaired} ones (CycleGAN~\cite{key2}, CUT~\cite{key3}) do not impose an a priori style/content separation as NST does. They instead rely on a \emph{tension} between two complementary forces: (i) a constraint that \emph{preserves} scene information (content), and (ii) a \emph{generative} constraint that pushes the produced image to adopt the appearance of the target domain (style).

In these three frameworks, appearance transformation is mainly induced by adversarial learning~\cite{goodfellow2020generative}, through a discriminator that distinguishes real images from the target domain and generated images. Content preservation is ensured by constraints specific to each method: cycle consistency (and often an $\ell_1$ penalty) in CycleGAN~\cite{key2}, a patch-based contrastive constraint (PatchNCE) combined with $\ell_1$ in CUT~\cite{key3}, and, in Pix2Pix~\cite{key1}, supervision by paired samples where an $\ell_1/\ell_2$ loss is sufficient to impose structural fidelity.

The interest of these approaches is that the boundary between ``what should change'' and ``what should remain'' is not fixed \emph{a priori}: it is determined \emph{by the task} through the balance between the adversarial term and the content constraints. For example, transforming a horse into a zebra mainly modifies a texture, whereas transforming a cat into a dog often requires more structural changes; similarly, simulating rain/snow acts mainly on appearance (Fig.~\ref{fig:example-style-transfer}). This flexibility enables more realistic transfer on real-world images.

A key element in this success is the use of a \emph{PatchGAN}-type discriminator. Introduced with Pix2Pix~\cite{key1}, PatchGAN evaluates the image in a \emph{local} way: the output corresponds to a grid of decisions, each neuron ``seeing'' only a restricted receptive field. This design makes the discriminator particularly sensitive to \emph{high-frequency details} (textures, micro-structures), which are precisely at the core of style.

However, PatchGAN must not be too deep: as depth increases, the receptive field widens and local predictors cease to be truly local (Fig.~\ref{fig:figure5}), which reduces the ability to constrain fine details. This observation converges with NST: Gatys \emph{et al.}~\cite{gatys2016image} also recommend extracting style from intermediate layers. It motivates our choice of a dedicated and \emph{shallow} mechanism (namely PatchGAN) for appearance: without an explicit style-capturing module, a very deep encoder naturally tends to favor content abstractions, at the risk of \emph{diluting} stylistic cues (as in purely predictive frameworks such as I-JEPA~\cite{assran2023self}).

\subsection{Transfer heuristics for robust disentanglement}
\label{subsec:disentanglement}

Our methodology exploits the synergy between unpaired style transfer (CUT~\cite{key3}). This approach offers a decisive advantage over NST-type methods~\cite{gatys2016image,ruta2023aladin}: instead of assuming a perfect orthogonality between style and content, it learns to modify appearance through adversarial constraints while maximizing structural similarity (contrastive loss). This flexibility makes it possible to model realistic and continuous styles (fog, rain) rather than simple artistic texture overlays. It is on this ability to selectively perturb appearance that our self-supervised learning relies. In the following, we further detail our methodology.

\section{Methodology}
\label{sec:methodology}

We aim at learning \emph{without annotations} two complementary representations
(\hyperref[fig:two-brains-architecture]{Fig.~\ref{fig:two-brains-architecture}}):
(i) a \emph{content} representation, robust to appearance variations and transferable to semantic tasks
(classification/detection), and (ii) a \emph{style} representation, dedicated to appearance signatures
(textures, spectrum, diffusion) and formulated in a \emph{predictable} way so that it can be reused when appearance is decisive
(e.g., fine weather characterization, histopathology).

\begin{figure}[!t]
    \centering
    \includegraphics[width=1.0\columnwidth]{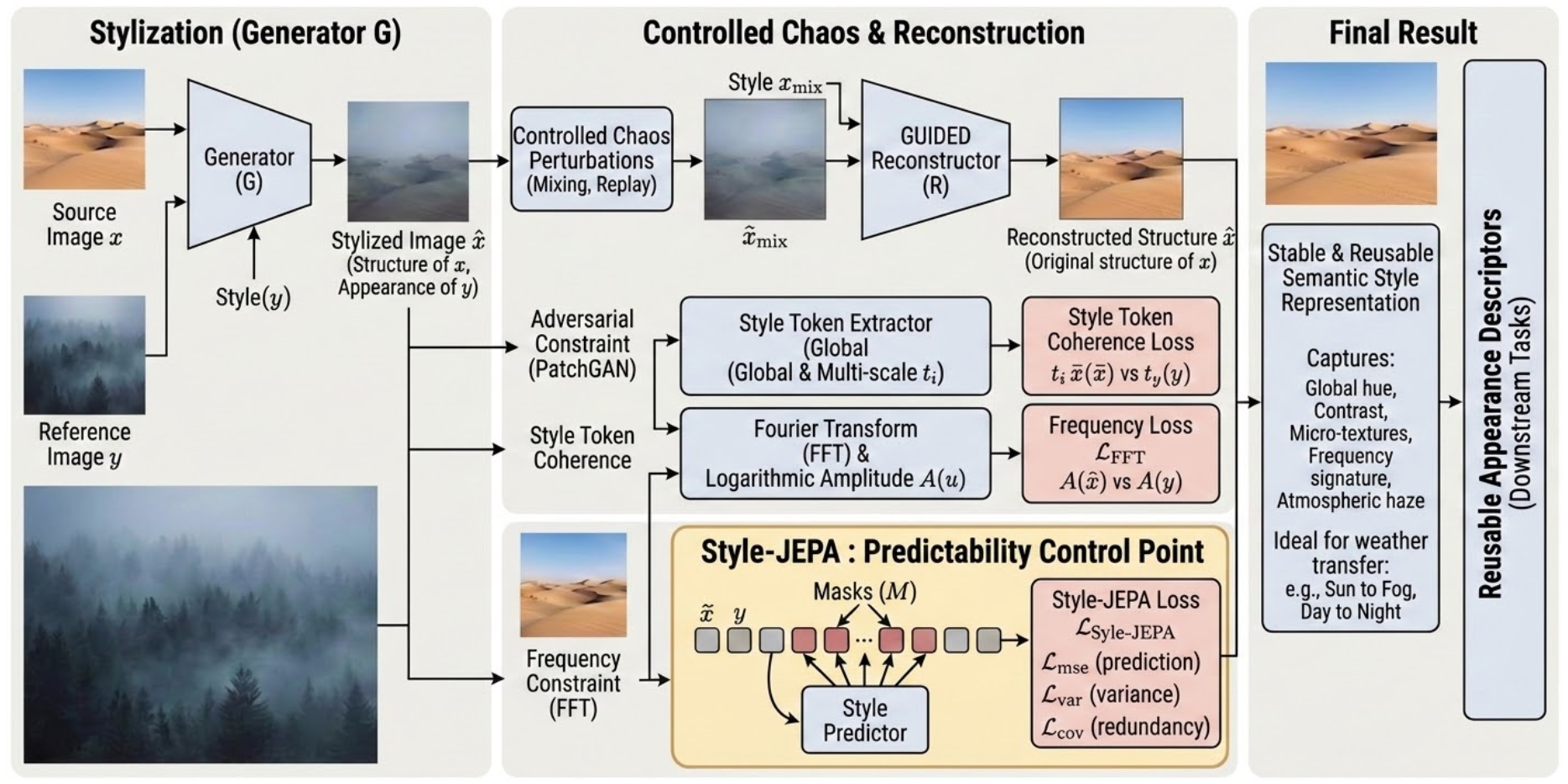}
    \caption{Semantic learning of style.
    (A) An image $x$ is stylized toward a target pseudo-domain under adversarial and frequency constraints.
    (B) Guided reconstruction and Style-JEPA filter contingent details and encourage stable and reusable style tokens.}
    \label{fig:style-learning-process}
\end{figure}

\begin{figure}[!t]
    \centering
    \includegraphics[width=0.9\columnwidth]{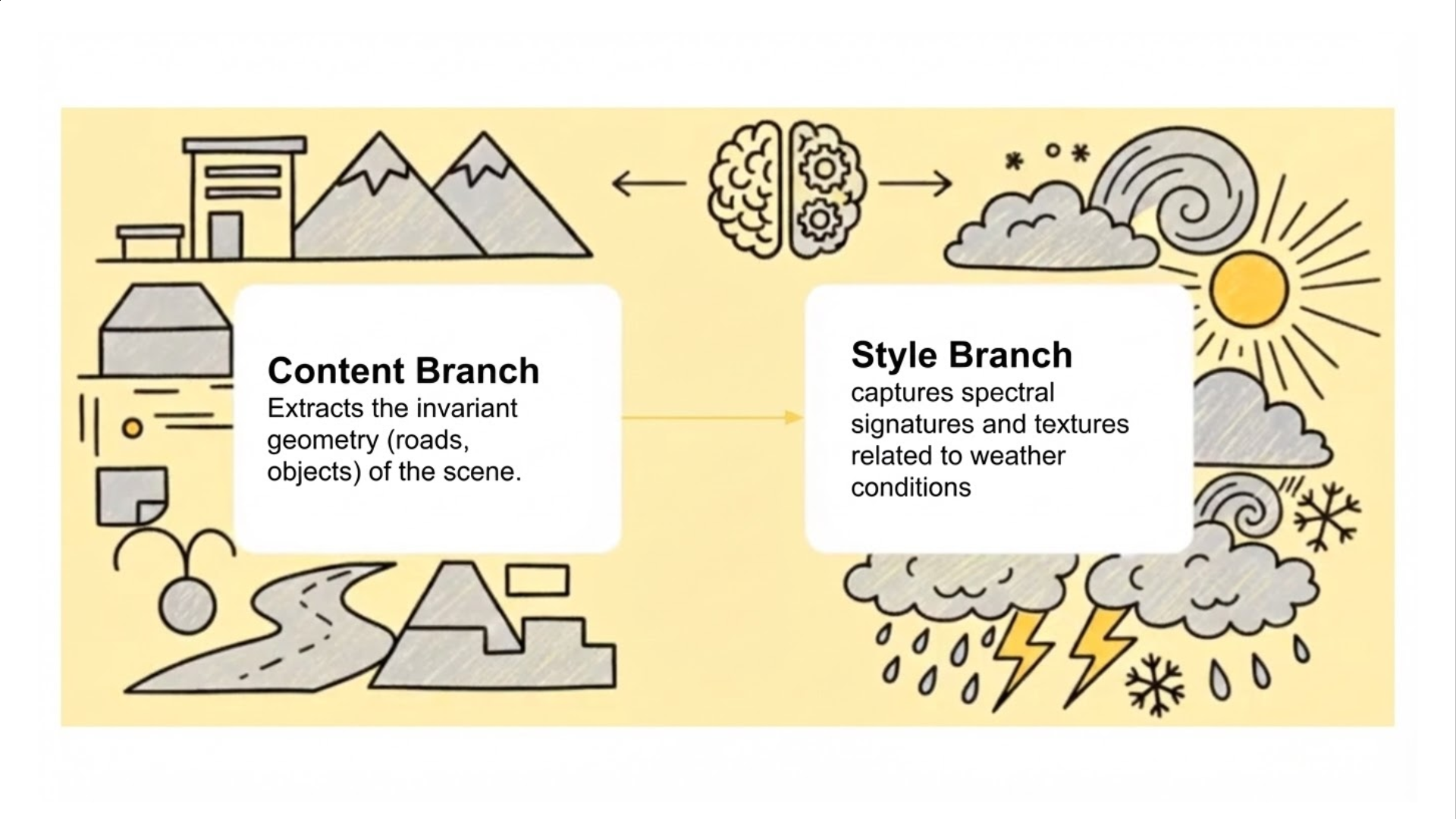}
    \caption{As in the human brain, where certain regions are specialized for certain tasks, ST-STORM is structured into two distinct poles:
    one stream dedicated to the extraction of invariant semantics (content), and one stream specialized in the predictive modeling of appearance (style).
    This separation makes it possible to apply to each modality an appropriate learning principle (invariance vs. predictability).}
    \label{fig:two-brains-architecture}
\end{figure}

For style, the difficulty is twofold: (a) preserving appearance information, which may be diluted by excessive invariance,
and (b) avoiding the memorization of contingent details if learning is dominated by pixel-to-pixel reconstruction.
ST-STORM relies on an organization into pseudo-domains, an explicit architectural disentanglement, and a joint semantic structuring
of the latent spaces (style and content) through adapted objectives (predictability for style, contrastive invariance for content).
In the following, we first present the partitioning and the training cycle, then the U-Net\,+\,SPADE disentanglement mechanism,
before detailing style and content learning, and finally the constraints that stabilize generation (\hyperref[fig:style-learning-process]{Fig.~\ref{fig:style-learning-process}}).

\subsection{Self-supervised partitioning into pseudo-domains and training cycle}
\label{subsec:global_cycle}

\begin{figure}[!t]
    \centering
    \includegraphics[width=0.9\columnwidth]{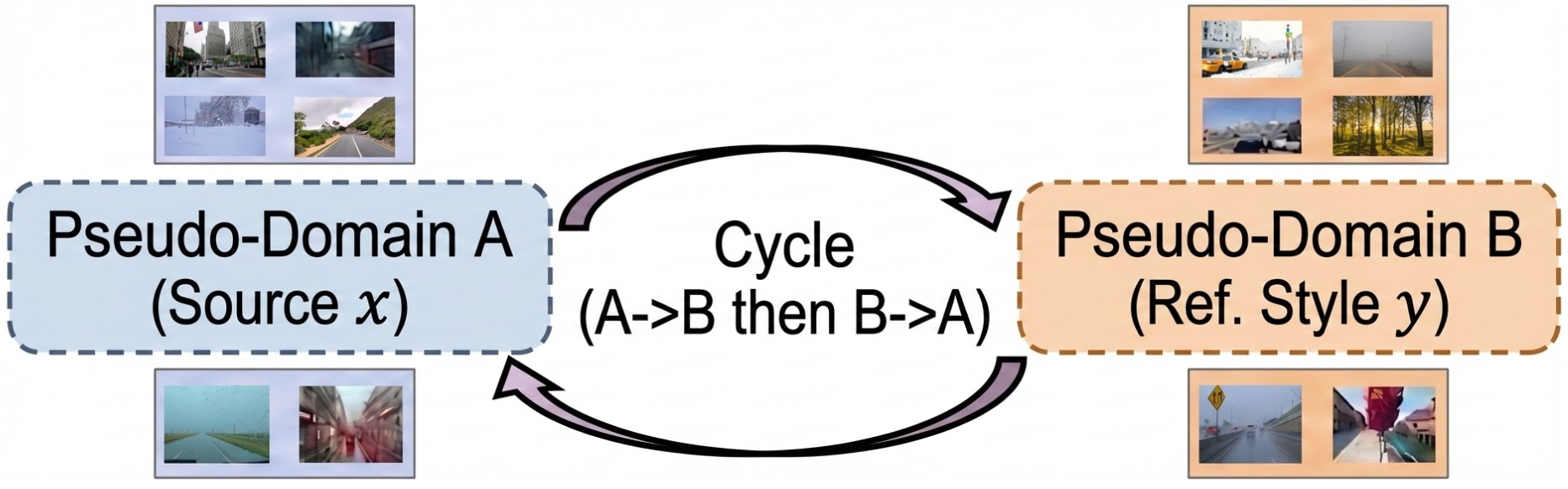}
    \caption{Partitioning into pseudo-domains and cyclic learning.
    The dataset is divided into arbitrary subsets in order to define self-supervised style transfer.
    Alternating transfer directions imposes systematic appearance perturbation, while stabilizing style targets over
    training windows.}
    \label{fig:leanrning-cycle}
\end{figure}

From an unlabeled set $\mathcal{D}$, we construct $K\ge 2$ subsets
$\{\mathcal{D}^{(k)}\}_{k=0}^{K-1}$ by random permutation followed by splitting.
These subsets define \emph{pseudo-domains} used only as appearance sources and targets during training
(\hyperref[fig:leanrning-cycle]{Fig.~\ref{fig:leanrning-cycle}}).

It is important to emphasize that our goal is not to assume the existence of perfectly separated styles at the dataset scale.
The partitioning mainly serves to \emph{perturb the appearance of an individual image} in a controlled (or learned) manner.
Thus, even if two pseudo-domains induce similar appearances on average, sampling $y$ from a different subset
than that of $x$ imposes a non-trivial constraint: for a given image, the reference $\mathrm{Style}(y)$ differs from the style of $x$ itself,
which produces a stylized view useful for learning. As illustrated by Figs.~\ref{fig:style-learning-process-melanoma} and~\ref{fig:style-learning-process-weather}.

Beyond this perturbation, the partitioning also plays a \emph{stabilization} role.
By maintaining relatively fixed source/target sets over training windows, it provides more coherent appearance targets
and reduces the risk of instability (or even collapse) that would arise if the appearance reference were drawn completely at random
and varied in a non-stationary way at each iteration.

Training is organized into \emph{rounds}. At round $r$, we select:
\begin{equation}
k_{\mathrm{src}}(r)=r \bmod K,\qquad
k_{\mathrm{tgt}}(r)=(r+1)\bmod K,
\end{equation}
then sample $x\sim \mathcal{D}^{(k_{\mathrm{src}})}$ and $y\sim \mathcal{D}^{(k_{\mathrm{tgt}})}$.
In the case $K=2$ (the most common one), this induces the alternation
$\mathcal{D}^{(0)}\!\to\!\mathcal{D}^{(1)}$ then $\mathcal{D}^{(1)}\!\to\!\mathcal{D}^{(0)}$.
This alternation avoids a unilateral definition of style, and forces the model to learn a notion of appearance
that is \emph{transferable} in both directions.

Within each round, optimization alternates three complementary phases:
(i) adversarial stylization (projection toward the target distribution),
(ii) style diversification (style bank, replay and/or spectral perturbations) in order to avoid a trivial correspondence,
and (iii) guided reconstruction that secures structural invariants.
This scheduler therefore explicitly determines \emph{when} appearance is deliberately perturbed and \emph{when} structure is stabilized,
which is central to obtaining reusable style without degrading content.

\subsection{Architectural disentanglement: explicit separation between structure and appearance}
\label{subsec:spade_role}

The central objective of ST-STORM is to separate two types of visual information that coexist in an image:
\begin{itemize}
    \item \emph{content}, that is, \emph{what is present in the scene} and \emph{where it is located}
    (shapes, contours, spatial layout, global geometry);
    \item \emph{style}, that is, \emph{how this scene appears}
    (texture, contrast, color, brightness, atmospheric diffusion, granularity, frequency signatures).
\end{itemize}

This separation must be \emph{operational}:
we want to be able to modify the appearance of an image without moving its structure.
In other words, if $x$ is a source image and $y$ an appearance reference image,
the model must be able to produce an image
that preserves the geometry of $x$ while adopting the style of $y$.

\begin{figure}[!t]
    \centering
    \includegraphics[width=0.9\columnwidth]{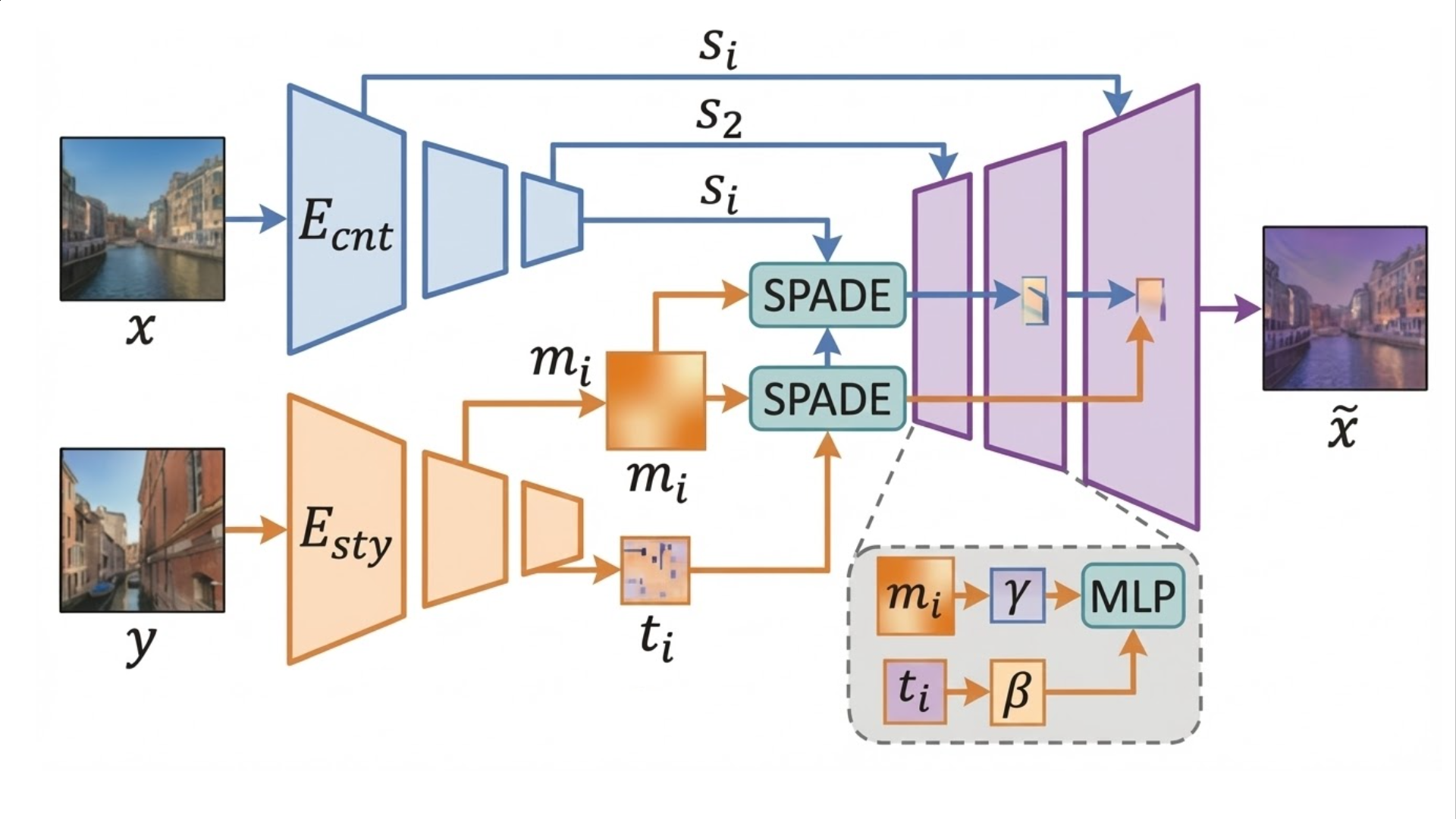}
    \caption{Architectural disentanglement of ST-STORM.
    The content path (U-Net) produces spatial representations $s_i$ transmitted through skip connections.
    The style path (pyramidal encoder) produces appearance maps $m_i$ and compact tokens $t_i$ together with a global token $t_G$.
    The SPADE blocks then fuse the two streams by modulating the decoder, which makes it possible to transform appearance without altering structure.}
    \label{fig:disenglement}
\end{figure}

To this end, ST-STORM uses a two-path architecture:
\begin{itemize}
    \item a \emph{content} path, based on a U-Net, which explicitly carries structure;
    \item a \emph{style} path, based on a pyramidal encoder, which extracts multi-scale appearance representations.
\end{itemize}
The fusion between these two streams is performed through SPADE blocks, which inject style not as a new geometry,
but as a \emph{modulation} of the decoder activations (Fig.~\ref{fig:disenglement}).

\paragraph*{General notations.}
We denote:
\begin{itemize}
    \item $x \in \mathbb{R}^{H \times W \times 3}$ the source image whose content we want to preserve;
    \item $y \in \mathbb{R}^{H \times W \times 3}$ the reference image from which we want to extract appearance;
    \item $G$ the main generator, which produces a stylized image from the content of $x$ and the style of $y$;
    \item $R$ the guided reconstruction operator used in phase B;
    \item $\tilde{x}$ the stylized image produced by the generator;
    \item $\hat{x}$ the image reconstructed during phase B.
\end{itemize}

The main style transfer is written as:
\begin{equation}
\tilde{x} = G\!\bigl(x;\,\mathrm{Style}(y)\bigr),
\end{equation}
where $\mathrm{Style}(y)$ denotes the set of style representations extracted from $y$.
In phase B, the model then performs a guided reconstruction:
\begin{equation}
\hat{x} = R\!\bigl(\tilde{x}_{\mathrm{mix}};\,\mathrm{Style}(x_{\mathrm{mix}})\bigr),
\end{equation}
where $x_{\mathrm{mix}}$ and $\tilde{x}_{\mathrm{mix}}$ denote samples possibly coming from a mixing or replay mechanism.

\subsubsection*{Content path}

Content is extracted from the source image $x$ by means of a U-Net encoder,
which we denote by $E_{\mathrm{cnt}}$.
This encoder produces a pyramid of multi-scale representations:
\begin{equation}
(s_1, s_2, s_3, s_4, s_5) = E_{\mathrm{cnt}}(x).
\end{equation}
Each tensor $s_i$ represents the scene at a given scale.
More precisely:
\begin{itemize}
    \item $s_i$ preserves a \emph{spatial organization};
    \item it mainly encodes \emph{structural} information:
    edges, shapes, object parts, relative arrangement of elements.
\end{itemize}

These tensors are passed to the decoder through \emph{skip connections}.
They play an essential role:
they provide the decoder with a direct path to reconstruct the geometry of the image,
which greatly reduces the incentive to encode this geometry in the style channel.

In other words, the representations $s_i$ answer the question:
\begin{center}
\emph{``What does the image contain, and where are the elements located?''}
\end{center}

\subsubsection*{Style path}

In parallel, a pyramidal style encoder, denoted $E_{\mathrm{sty}}$,
analyzes the reference image $y$ and extracts from it two types of complementary representations:
\begin{itemize}
    \item \emph{appearance maps} $m_i$;
    \item compact \emph{style tokens} $t_i$, together with a global token $t_G$.
\end{itemize}

Formally:
\begin{equation}
\bigl(m_1,\dots,m_5\bigr),\ \bigl(t_1,\dots,t_5\bigr),\ t_G = E_{\mathrm{sty}}(y).
\end{equation}

\paragraph*{Local appearance maps.}
For each scale $i \in \{1,\dots,5\}$,
the tensor
\[
m_i \in \mathbb{R}^{H_i \times W_i \times C_i}
\]
is a style feature map,
where $H_i$ and $W_i$ are the spatial dimensions at scale $i$,
and $C_i$ the number of channels.
This map preserves a spatial structure and describes \emph{where} certain visual properties appear:
brighter areas, denser haze, more marked texture, more reflective regions, etc.

Thus, $m_i$ carries a \emph{spatially localized} style.
For example, in a foggy scene, fog may be more intense in some regions than in others;
a map $m_i$ can represent this spatial variation.

\paragraph*{Multi-scale style tokens}
At each scale $i$, the map $m_i$ is condensed into a compact style vector.
This operation is performed in two steps:
\begin{enumerate}
    \item a spatial aggregation, denoted $\mathrm{Pool}(\cdot)$;
    \item a learned linear projection.
\end{enumerate}

We define:
\begin{equation}
u_i = \mathrm{Pool}(m_i),
\end{equation}
where
\[
u_i \in \mathbb{R}^{C_i}
\]
is a vector summarizing the map $m_i$ at scale $i$.
The function $\mathrm{Pool}$ denotes here a spatial compression operation
(for example average pooling or an equivalent aggregation over the spatial dimensions).

This summary is then projected into a common latent space of dimension $d_t$:
\begin{equation}
t_i = W_i\,u_i + b_i,
\end{equation}
where:
\begin{itemize}
    \item $W_i \in \mathbb{R}^{d_t \times C_i}$ is a learned projection matrix;
    \item $b_i \in \mathbb{R}^{d_t}$ is a learned bias;
    \item $t_i \in \mathbb{R}^{d_t}$ is the style token at scale $i$.
\end{itemize}

Each token $t_i$ therefore summarizes appearance at a particular scale:
fine textures, intermediate structures, or more global patterns depending on the considered depth.

\paragraph*{Global style token}
In addition to the multi-scale tokens $t_i$,
the model builds a global token $t_G$ from the last pyramidal map $m_5$.
This token is intended to summarize the global appearance of the whole scene.

We first define:
\begin{equation}
u_G = \mathrm{GAP}(m_5),
\end{equation}
where $\mathrm{GAP}$ denotes a \emph{Global Average Pooling},
and where
\[
u_G \in \mathbb{R}^{C_5}
\]
is a global vector describing the average statistics of the last style map.

The global token is then obtained by projection:
\begin{equation}
t_G = W_G\,u_G + b_G,
\end{equation}
where:
\begin{itemize}
    \item $W_G \in \mathbb{R}^{d_t \times C_5}$ is a learned matrix;
    \item $b_G \in \mathbb{R}^{d_t}$ is a learned bias;
    \item $t_G \in \mathbb{R}^{d_t}$ is the global style token.
\end{itemize}

The two types of representations do not play the same role:
\begin{itemize}
    \item the maps $m_i$ describe \emph{how style varies locally in space};
    \item the tokens $t_i$ and $t_G$ summarize \emph{more compact and more global statistics of appearance}.
\end{itemize}

Their function can be summarized as follows:
\begin{itemize}
    \item $m_i$ answers \emph{``where and how does appearance vary in the image?''};
    \item $t_i$ and $t_G$ answer \emph{``what is the dominant style at this scale or globally?''}
\end{itemize}

The fusion between content and style is performed through SPADE blocks
(\emph{Spatially-Adaptive Denormalization})~\cite{park2019semanticimagesynthesisspatiallyadaptive}.
The main idea is the following:
\begin{quote}
structure must not be generated by style;
style must only \emph{modulate} a structure that is already present.
\end{quote}

At scale $i$, let
\[
h_i \in \mathbb{R}^{H_i \times W_i \times C_i'}
\]
be the decoder activation tensor before modulation,
where $C_i'$ denotes the number of decoder channels at this scale.
We first apply an instance-wise normalization:
\begin{equation}
\hat{h}_i = \mathrm{IN}(h_i),
\end{equation}
where $\mathrm{IN}(\cdot)$ denotes an \emph{Instance Normalization}.

The SPADE block then produces an affine modulation:
\begin{equation}
\mathrm{SPADE}(h_i; m_i, t_i) = \alpha_i \odot \hat{h}_i + \delta_i,
\label{eq:spade_compact_rewrite}
\end{equation}
where:
\begin{itemize}
    \item $\alpha_i$ is a multiplicative gain;
    \item $\delta_i$ is an additive bias;
    \item $\odot$ denotes the Hadamard product, that is, the element-wise product.
\end{itemize}

The modulation parameters come from two sources:
\begin{itemize}
    \item a \emph{spatial} component, derived from the map $m_i$;
    \item a \emph{global} component, derived from the token $t_i$.
\end{itemize}

We write:
\begin{equation}
\alpha_i = 1 + \gamma_m(m_i) + \gamma_t(t_i),
\end{equation}
\begin{equation}
\delta_i = \beta_m(m_i) + \beta_t(t_i),
\end{equation}
where:
\begin{itemize}
    \item $\gamma_m(\cdot)$ and $\beta_m(\cdot)$ are learned functions applied to the map $m_i$
    (generally implemented by small convolutional networks);
    \item $\gamma_t(\cdot)$ and $\beta_t(\cdot)$ are learned functions applied to the token $t_i$
    (generally implemented by small linear layers);
    \item the constant term $1$ in $\alpha_i$ makes it possible to preserve a local identity when modulation is weak.
\end{itemize}

This formulation is important:
\begin{itemize}
    \item the main spatial structure continues to be carried by the decoder activations and the skips $s_i$;
    \item style acts as a \emph{reweighting} or a \emph{shift} of channels,
    and not as an autonomous geometric signal.
\end{itemize}

\subsubsection*{Why this architecture favors disentanglement}

This architectural factorization makes content leakage into the style branch less advantageous for the model.
More precisely:

\paragraph*{(i)}
The skips $s_i$ provide the decoder with a direct and stable path to reconstruct geometry.
It is therefore simpler for the model to use the representations $s_i$ for structure
than to try to re-encode this structure into style.

\paragraph*{(ii)}
Style is obtained from $y$, then applied to the structure of $x$.
Since the geometries of $x$ and $y$ generally differ,
directly encoding the structure of $y$ into style would be of little use,
because this structure would not naturally align with that of $x$.

\paragraph*{(iii)}
SPADE blocks mainly modify the statistics of the decoder activations
(gain and bias per channel),
which favors appearance changes
(texture, contrast, diffusion, colors),
without providing a natural mechanism to move objects or change their arrangement.

\paragraph*{(iv)}
Content consistency, guided reconstruction, and predictability losses
penalize structural drifts.
Thus, even if style could in theory memorize content,
this becomes poorly compatible with the learning objectives.

In the end, the decoder receives two complementary streams:
\begin{itemize}
    \item content via the skips $s_i$, which answers the \emph{where/what};
    \item style via the maps $m_i$ and the tokens $t_i$, which answers the \emph{how}.
\end{itemize}

\subsection{Semantic learning of style}
\label{subsec:style_semantics}

We seek a style representation that captures stable and reusable appearance factors
(global hue, contrast, micro-textures, frequency signature, atmospheric veil),
while avoiding the memorization of contingent details and the leakage of geometric content into style
(\hyperref[fig:style-learning-process]{Fig.~\ref{fig:style-learning-process}}).

Style learning relies on two complementary operations:
\begin{enumerate}
    \item \emph{stylizing} a source image $x$ using the style of a reference image $y$;
    \item then \emph{reconstructing} a coherent structure from a perturbed stylized view.
\end{enumerate}

At a given iteration, the generator first produces a stylized image
\begin{equation}
\tilde{x}=G\!\bigl(x;\mathrm{Style}(y)\bigr),
\end{equation}
where $x$ provides the structure and $y$ provides the appearance signals.
The model then applies a guided reconstruction:
\begin{equation}
\hat{x}=R\!\bigl(\tilde{x}_{\mathrm{mix}};\mathrm{Style}(x_{\mathrm{mix}})\bigr),
\end{equation}
where $x_{\mathrm{mix}}$ is a reference image for reconstruction
and $\tilde{x}_{\mathrm{mix}}$ its stylized or mixed version coming from a \emph{mix} or \emph{replay} mechanism.
This alternation simultaneously imposes two properties:
\begin{itemize}
    \item style must be rich enough to visibly modify appearance;
    \item structure must remain recoverable, which discourages encoding geometry in the style tokens.
\end{itemize}

\paragraph*{Adversarial constraint}
The realism of the stylized image $\tilde{x}$ is enforced by a PatchGAN discriminator,
denoted by $D$,
which locally judges whether an image belongs to the target pseudo-domain or not.
With a hinge-loss, we write:
\begin{align}
\mathcal{L}_{D}^{\mathrm{hinge}}
&=
\mathbb{E}_{y}\!\left[\max(0,1-D(y))\right]
+
\mathbb{E}_{\tilde{x}}\!\left[\max(0,1+D(\tilde{x}))\right],\\
\mathcal{L}_{G}^{\mathrm{adv}}
&=
-\mathbb{E}_{\tilde{x}}\!\left[D(\tilde{x})\right].
\end{align}
The local judgment of PatchGAN is well suited to styles dominated by textures,
high-frequency signatures, and local irregularities.
An LSGAN variant can also be used.

\paragraph*{Style token consistency constraint}
Beyond visual realism, we want the stylized image $\tilde{x}$ to actually adopt the style of the reference $y$.
To do so, we directly compare the style tokens extracted from $\tilde{x}$ with those extracted from $y$:
\begin{equation}
\mathcal{L}_{\mathrm{sty\text{-}tok}}
=
\|t_G(\tilde{x})-t_G(y)\|_1
+
\sum_{i=1}^{5}\|t_i(\tilde{x})-t_i(y)\|_1.
\end{equation}
Here:
\begin{itemize}
    \item $t_i(\tilde{x})$ denotes the style token at scale $i$ extracted from the stylized image;
    \item $t_i(y)$ denotes the corresponding token extracted from the reference image;
    \item $t_G(\tilde{x})$ and $t_G(y)$ are the global tokens.
\end{itemize}
This constraint aligns style in a compact latent space,
which is more robust than a pixel-to-pixel comparison.

To avoid a trivial correspondence between an image $x$ and a single reference $y$,
the model is subjected to \emph{controlled chaos}:
style bank, reference mixing, and replay of stylized examples.
The role of this device is to force the style tokens
to capture transverse regularities of the target pseudo-domain,
rather than memorizing accidental correlations between a source image and a single reference.

Additional frequency constraints can also be explored,
but they do not constitute the core of the method.
We consider them as optional refinements,
studied separately in the ablation experiments.

Guided reconstruction turns this chaos into a useful signal.
It imposes that, despite a strongly perturbed appearance,
structure remains recoverable.
Coupled with content constraints, for example PatchNCE,
it discourages encoding geometry in the style channel:
such leakage would be unstable under mix/replay and would harm reconstruction (Figs.~\ref{fig:style-learning-process-melanoma} and~\ref{fig:style-learning-process-weather}).

\begin{figure}[!t]
    \centering
    \includegraphics[width=1.0\columnwidth]{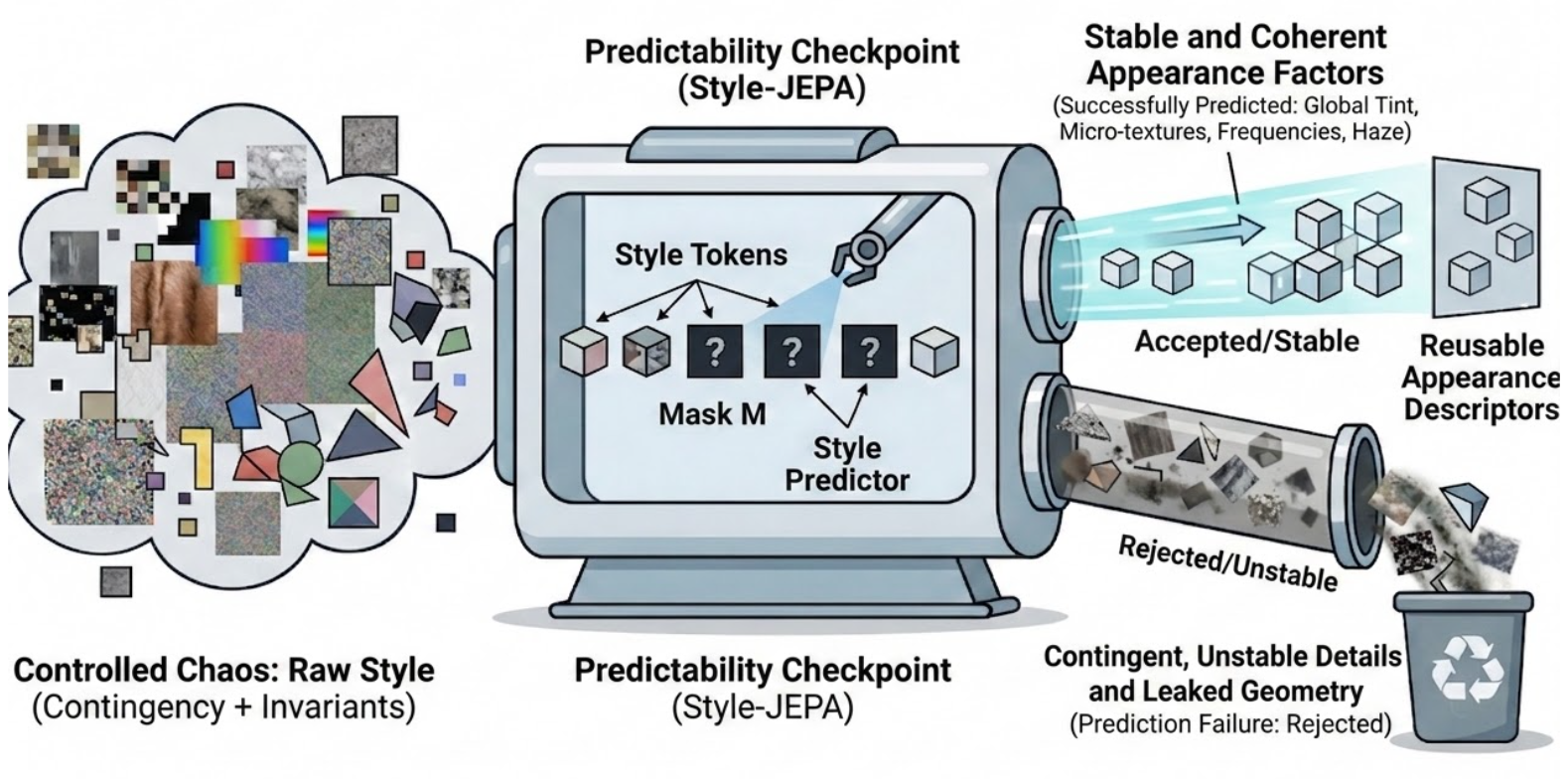}
    \caption{The predictability constraint acts as a filter that removes contingent and unpredictable details, while preserving coherent and reusable appearance factors.}
    \label{fig:style-JEPA}
\end{figure}

\begin{figure}[!t]
    \centering
    \includegraphics[width=1.0\columnwidth]{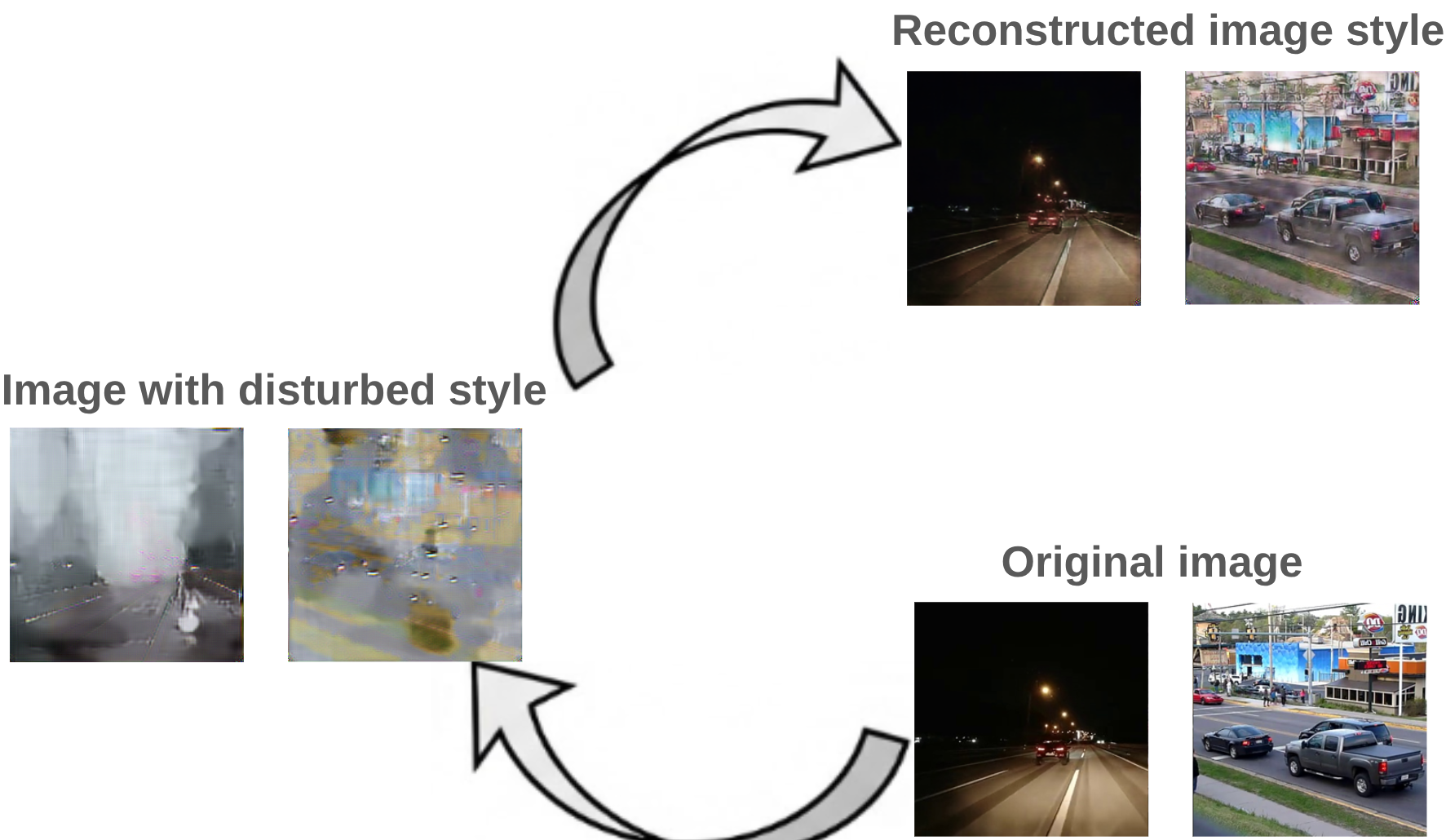}
    \caption{Controlled and learned perturbation of \emph{style} on an image from the ISIC 2024 Challenge dataset.
    The model mainly modifies appearance signatures, then learns to reconstruct the original appearance under a predictability constraint,
    in order to obtain semantic and reusable style tokens for downstream tasks.}
    \label{fig:style-learning-process-melanoma}
\end{figure}

\begin{figure}[!t]
    \centering
    \includegraphics[width=1.0\columnwidth]{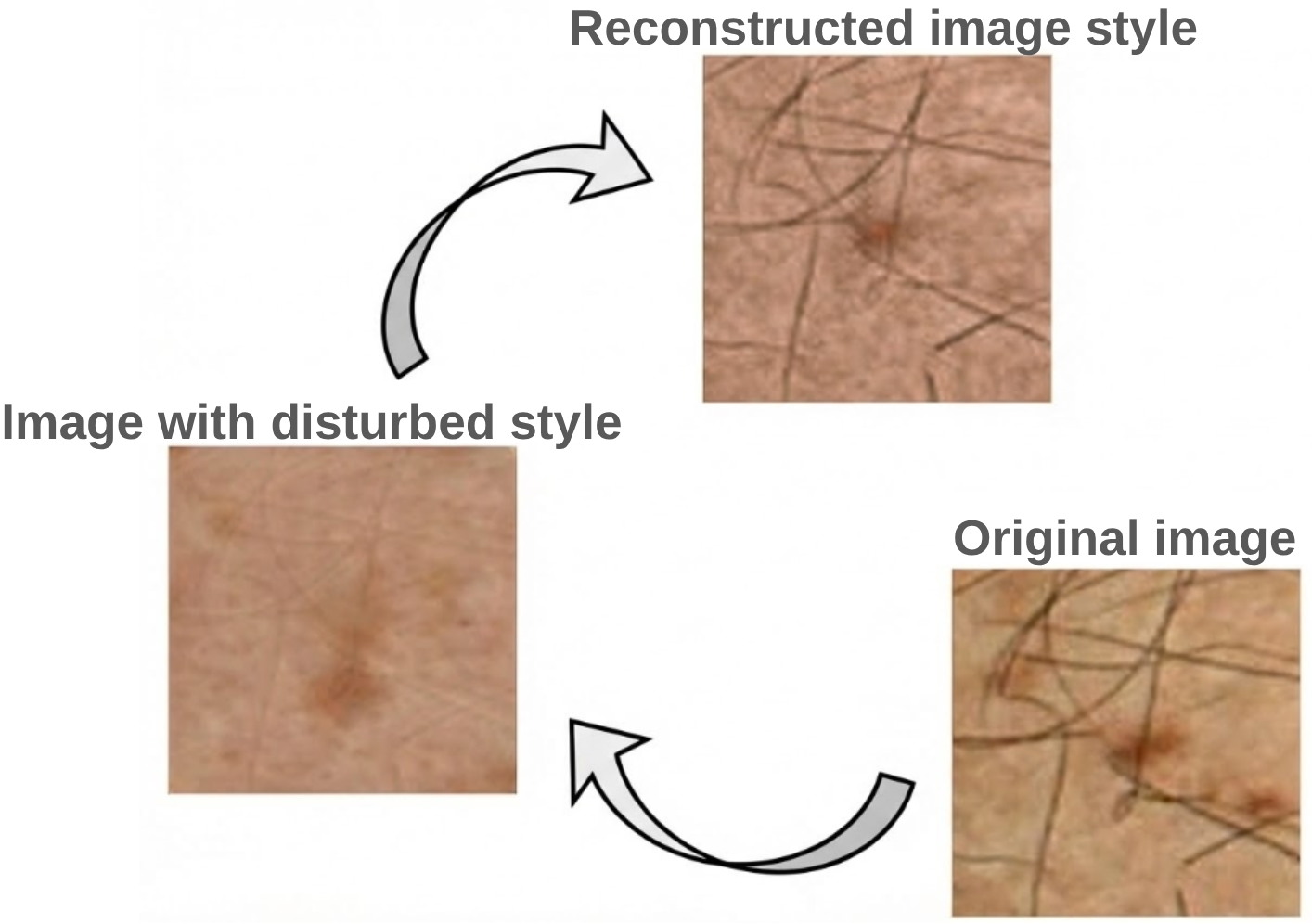}
    \caption{Controlled and learned perturbation of \emph{style} on weather images; the model intensifies appearance perturbations during training.
    The model mainly modifies appearance signatures, then learns to reconstruct the original appearance under a predictability constraint for downstream weather detection tasks.}
    \label{fig:style-learning-process-weather}
\end{figure}

\paragraph*{Style-JEPA: filtering contingency through predictability}
Finally, we impose a predictability constraint on style tokens so as to retain only coherent and stable components (Fig.~\ref{fig:style-JEPA}).

The model predicts a masked part of a sequence of style tokens from the visible context.
The total objective is written as:
\begin{equation}
\mathcal{L}_{\mathrm{StyleJEPA}}
=
\mathcal{L}_{\mathrm{mse}}
+
\lambda_{\mathrm{var}}\mathcal{L}_{\mathrm{var}}
+
\lambda_{\mathrm{cov}}\mathcal{L}_{\mathrm{cov}},
\label{eq:stylejepa_total_rewrite}
\end{equation}
where:
\begin{itemize}
    \item $\mathcal{L}_{\mathrm{mse}}$ measures the prediction error of masked tokens;
    \item $\mathcal{L}_{\mathrm{var}}$ penalizes variance collapse;
    \item $\mathcal{L}_{\mathrm{cov}}$ penalizes redundancy between dimensions;
    \item $\lambda_{\mathrm{var}}$ and $\lambda_{\mathrm{cov}}$ are scalar weights.
\end{itemize}

The main prediction term is written as:
\begin{equation}
\mathcal{L}_{\mathrm{mse}}
=
\frac{1}{|\mathcal{M}|}
\sum_{(b,s)\in\mathcal{M}}
\left\|
\hat{\mathbf{T}}_{b,s}-\mathrm{sg}(\mathbf{T}_{t,b,s})
\right\|_2^2,
\label{eq:stylejepa_mse_rewrite}
\end{equation}
where:
\begin{itemize}
    \item $\mathcal{M}$ is the set of masked positions;
    \item $b$ indexes the batch element;
    \item $s$ indexes the token position or scale in the sequence;
    \item $\hat{\mathbf{T}}_{b,s}$ is the token predicted by the predictor;
    \item $\mathbf{T}_{t,b,s}$ is the target token provided by the teacher network;
    \item $\mathrm{sg}(\cdot)$ denotes stop-gradient.
\end{itemize}

Predictability here acts as a filter:
contingent, unpredictable, and unstable details are penalized,
while coherent appearance factors are preserved.
Thus, style tokens become not only useful for generation,
but also \emph{reusable downstream} as appearance descriptors.

\subsection{Semantic learning of content through MoCo with stylized positives}
\label{subsec:content_moco}

\begin{figure}[!t]
    \centering
    \includegraphics[width=0.9\columnwidth]{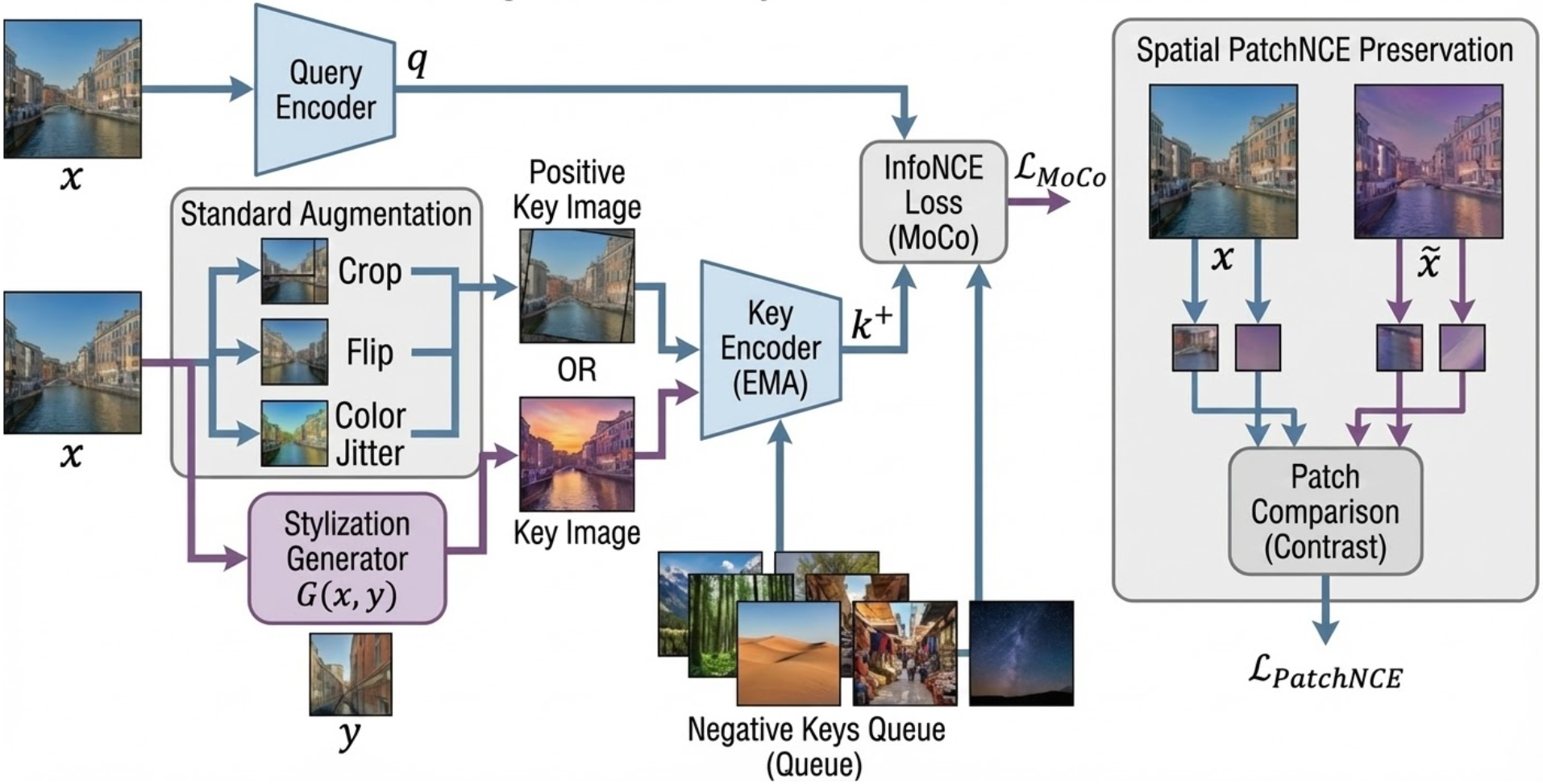}
    \caption{Semantic learning of content with MoCo.
    The main difference from classical MoCo is the introduction of stylized views as positives,
    which forces the content representation to become invariant to appearance perturbations.}
    \label{fig:content-learning-process}
\end{figure}

The content representation must remain stable under appearance variations,
including those induced by stylization
(\hyperref[fig:content-learning-process]{Fig.~\ref{fig:content-learning-process}}).
The objective here is no longer to represent \emph{how the image appears},
but \emph{what it contains} independently of changes in texture, contrast, or color.

To this end, we use MoCo-style contrastive learning.
The principle is the following:
two views of the same content should produce similar representations,
even if their appearance differs strongly.
In ST-STORM, these two views can be obtained:
\begin{itemize}
    \item either through standard augmentations;
    \item or through controlled stylization;
    \item or through a combination of both.
\end{itemize}

We denote:
\begin{itemize}
    \item $x^{(q)}$ the query view;
    \item $x^{(k)}$ the key view;
    \item $f_q$ the query encoder;
    \item $f_k$ the key encoder;
    \item $q=f_q(x^{(q)}) \in \mathbb{R}^{d_c}$ the query embedding;
    \item $k=f_k(x^{(k)}) \in \mathbb{R}^{d_c}$ the associated positive embedding;
    \item $\mathcal{Q}=\{k_i\}_{i=1}^{K}$ the queue of $K$ negative embeddings.
\end{itemize}

The InfoNCE contrastive loss is written as:
\begin{equation}
\mathcal{L}_{\mathrm{MoCo}}
=
-\log
\frac{\exp(\langle q,k\rangle/\tau)}
{\exp(\langle q,k\rangle/\tau)+\sum_{i=1}^{K}\exp(\langle q,k_i\rangle/\tau)},
\label{eq:moco_safe_rewrite}
\end{equation}
where:
\begin{itemize}
    \item $\langle q,k\rangle$ denotes the dot product between two embeddings;
    \item $\tau>0$ is a temperature;
    \item $k_i$ denotes the $i$-th negative present in the queue.
\end{itemize}

This loss encourages:
\begin{itemize}
    \item proximity between two views of the same content;
    \item separation between views coming from different contents.
\end{itemize}

The key encoder is updated by exponential moving average:
\begin{equation}
\theta_k \leftarrow m\,\theta_k + (1-m)\,\theta_q,
\end{equation}
where:
\begin{itemize}
    \item $\theta_q$ denotes the parameters of the query encoder;
    \item $\theta_k$ denotes the parameters of the key encoder;
    \item $m \in [0,1)$ is the momentum coefficient.
\end{itemize}

The main interest of this strategy in ST-STORM
is to introduce \emph{stylized positives}.
Thus, the content branch explicitly learns
that a change in appearance must not be interpreted as a change in content.
It therefore becomes more robust to style perturbations,
and provides a latent space better suited to tasks where object structure dominates.

\subsection{Stabilization and content preservation in generation}
\label{subsec:phaseB}

The stability of disentanglement is reinforced by guided reconstruction from stylized views coming from mix/replay:
\begin{equation}
\hat{x}=R\!\bigl(\tilde{x}_{\text{mix}};\,\mathrm{Style}(x_{\text{mix}})\bigr).
\end{equation}
Here:
\begin{itemize}
    \item $\tilde{x}_{\text{mix}}$ is a stylized or perturbed image;
    \item $x_{\text{mix}}$ provides the reference style used to guide reconstruction.
\end{itemize}

This step imposes that, even after a strong appearance perturbation,
structure must remain recoverable.
It therefore plays a disentanglement stabilization role:
style may vary, but content must not be lost.

In addition, a PatchNCE constraint locks spatial correspondences between the source image and the stylized image.
Let $\phi_l(\cdot)$ be the features extracted at layer $l$ of a given network,
and let $p$ be a spatial position in the feature map.
We define:
\begin{equation}
q_{l,p}=\phi_l(x)_p,\qquad k^{+}_{l,p}=\phi_l(\tilde{x})_p,
\end{equation}
where:
\begin{itemize}
    \item $q_{l,p}$ is the feature vector of the source image at layer $l$ and position $p$;
    \item $k^{+}_{l,p}$ is the corresponding positive vector in the stylized image;
    \item $\mathcal{N}_{l,p}$ denotes the set of negatives for this layer and position.
\end{itemize}

The PatchNCE loss is written as:
\begin{align}
\mathcal{L}_{\mathrm{PatchNCE}}
&=
\sum_{l}\mathbb{E}_{p}\Bigg[
-\log
\frac{\exp(\langle q_{l,p},k^{+}_{l,p}\rangle/\tau)}
{\sum_{k\in\{k^{+}_{l,p}\}\cup\mathcal{N}_{l,p}}
\exp(\langle q_{l,p},k\rangle/\tau)}
\Bigg].
\end{align}

This constraint pushes the model to preserve, locally, content correspondences between the source image and the transformed image.
It reduces structural drifts
and strengthens the role of content representations in the generative process.

\subsection{How style tokens are used in downstream tasks}
\label{subsec:downstream_tokens}

The style tokens extracted by $E_{\mathrm{sty}}$ are not only used to guide generation.
They also constitute a compact reusable representation for classification tasks, attribute prediction,
or fusion with semantic embeddings.

\paragraph*{Set of available tokens.}
From an image $x$, the style encoder provides:
\[
(t_1,\dots,t_5), \qquad t_G,
\]
where each vector belongs to $\mathbb{R}^{d_t}$.
Several downstream representations can then be constructed depending on the task.

\paragraph*{Global token only.}
The simplest representation consists in using only the global token:
\begin{equation}
f_{\mathrm{sty}} = t_G,
\end{equation}
where $f_{\mathrm{sty}}$ denotes the style embedding used downstream.
This option is compact and captures the dominant appearance of the image.

\paragraph*{Multi-scale concatenation.}
One may also concatenate the global token and the local tokens:
\begin{equation}
f_{\mathrm{sty}} =
[t_G \,\|\, t_5 \,\|\, t_4 \,\|\, t_3 \,\|\, t_2 \,\|\, t_1],
\end{equation}
where $\|$ denotes concatenation.
This representation explicitly preserves style information at multiple scales.

\paragraph*{Multi-scale average.}
Another possibility is to aggregate the tokens by averaging:
\begin{equation}
f_{\mathrm{sty}} =
\frac{1}{6}\left(t_G + \sum_{i=1}^{5} t_i\right).
\end{equation}
This yields a more compact embedding, less sensitive to a particular scale.

\paragraph*{Weighted aggregation.}
Finally, one may use a weighted combination:
\begin{equation}
f_{\mathrm{sty}} =
w_G\,t_G + \sum_{i=1}^{5} w_i\,t_i,
\end{equation}
where the scalar weights
\[
w_G,\ w_1,\dots,w_5 \in \mathbb{R}
\]
are either fixed manually or learned.

\paragraph*{Use in supervised classification.}
Once the style embedding $f_{\mathrm{sty}}$ has been constructed,
it is passed to a supervised head $H$:
\begin{equation}
\hat{y} = H(f_{\mathrm{sty}}),
\end{equation}
where:
\begin{itemize}
    \item $f_{\mathrm{sty}} \in \mathbb{R}^{d_f}$ is the final embedding used for the task;
    \item $H$ is a classification or multi-task prediction head;
    \item $\hat{y}$ denotes the prediction produced by the model.
\end{itemize}

In the multi-task case, $H$ contains one classifier per attribute
(for example weather type, intensity, visibility, ground state).
In the single-task case, it is a standard classifier over the full set of classes.

\paragraph*{Intuition for downstream use}
The interest of style tokens is that they constitute a \emph{compact yet structured} representation of appearance.
They allow the downstream model to work directly on appearance descriptors,
instead of having to relearn this information from the raw image.
The multi-scale tokens also make it possible to isolate fine or global appearance cues depending on the task.

\subsection{Adaptive fusion of style and semantic content representations}
\label{subsec:fusion_style_semantic}

In some tasks, especially large-scale object recognition,
purely stylistic representations may be insufficient when used alone.
They efficiently capture appearance information
(textures, contrasts, local signatures, global statistics),
but these signals are not always the most discriminative for identifying an object category.
Conversely, semantic representations coming from a content-dedicated backbone
are generally better aligned with the geometry, structure, and organization of object parts,
but may lose some of the fine appearance cues.
To exploit the complementarity of these two sources,
we introduce a lightweight adaptive fusion between style embeddings and semantic embeddings.

Let $x$ be an input image.
The stylistic branch produces a global vector
\[
f_{\mathrm{sty}} \in \mathbb{R}^{d_{\mathrm{sty}}},
\]
obtained from the aggregated style tokens.
In parallel, the semantic branch produces a global vector
\[
f_{\mathrm{sem}} \in \mathbb{R}^{d_{\mathrm{sem}}},
\]
extracted from the semantic backbone after spatial aggregation.

Since these two representations come from different modules
and do not necessarily have the same distribution or dimension,
we first apply an independent normalization to each of them:
\[
\hat f_{\mathrm{sty}} = \mathrm{LN}(f_{\mathrm{sty}}),
\qquad
\hat f_{\mathrm{sem}} = \mathrm{LN}(f_{\mathrm{sem}}),
\]
where $\mathrm{LN}$ denotes a \emph{Layer Normalization}.

The two normalized vectors are then projected into a common latent space of dimension $d_f$:
\[
z_{\mathrm{sty}} = P_{\mathrm{sty}}(\hat f_{\mathrm{sty}}),
\qquad
z_{\mathrm{sem}} = P_{\mathrm{sem}}(\hat f_{\mathrm{sem}}),
\]
where:
\begin{itemize}
    \item $P_{\mathrm{sty}} : \mathbb{R}^{d_{\mathrm{sty}}} \rightarrow \mathbb{R}^{d_f}$ is a learned projection for style;
    \item $P_{\mathrm{sem}} : \mathbb{R}^{d_{\mathrm{sem}}} \rightarrow \mathbb{R}^{d_f}$ is a learned projection for semantic content.
\end{itemize}

Rather than abruptly concatenating the two vectors,
we then learn a lightweight \emph{vector gating} mechanism,
which allows the model to weight, dimension by dimension,
the relative contribution of each branch.
We first concatenate the two projected representations:
\[
z_{\mathrm{cat}} = [z_{\mathrm{sem}} \,\|\, z_{\mathrm{sty}}] \in \mathbb{R}^{2d_f},
\]
then compute a gate vector
\[
g = \sigma\!\bigl(W_g\, z_{\mathrm{cat}} + b_g\bigr) \in \mathbb{R}^{d_f},
\]
where:
\begin{itemize}
    \item $W_g \in \mathbb{R}^{d_f \times 2d_f}$ is a learned matrix;
    \item $b_g \in \mathbb{R}^{d_f}$ is a learned bias;
    \item $\sigma(\cdot)$ is the sigmoid applied component-wise.
\end{itemize}

Each component $g_k \in [0,1]$ controls, for dimension $k$,
the share of information coming from semantic content
relative to that coming from style.

The fused representation is then defined by
\[
z_{\mathrm{fus}} = g \odot z_{\mathrm{sem}} + (1-g) \odot z_{\mathrm{sty}},
\]
where $\odot$ denotes the Hadamard product.
Thus, if one dimension of the gate vector is close to $1$,
the fused representation favors semantic information;
if it is close to $0$, it favors style information.

In order to stabilize the final embedding sent to the classifier,
we apply an additional normalization:
\[
\tilde z_{\mathrm{fus}} = \mathrm{LN}(z_{\mathrm{fus}}).
\]
The vector $\tilde z_{\mathrm{fus}}$ is then used as input to the supervised head:
\[
\hat y = H(\tilde z_{\mathrm{fus}}),
\]
where $H$ denotes the supervised head.

The complete mechanism can be summarized by:
\begin{equation}
\begin{aligned}
x
&\longrightarrow \bigl(f_{\mathrm{sty}},\,f_{\mathrm{sem}}\bigr)
\longrightarrow \bigl(\hat f_{\mathrm{sty}},\,\hat f_{\mathrm{sem}}\bigr)\\
&\longrightarrow \bigl(z_{\mathrm{sty}},\,z_{\mathrm{sem}}\bigr)
\longrightarrow g
\longrightarrow z_{\mathrm{fus}}
\longrightarrow \tilde z_{\mathrm{fus}}
\longrightarrow H
\longrightarrow \hat y.
\end{aligned}
\end{equation}

This fusion operates at the \emph{embedding} level,
which makes it possible to keep memory and computation costs moderate,
while allowing the model to learn
when it should rely more on semantic content
and when, on the contrary, style information is useful for the decision.
In other words, the fusion module does not oppose style and semantics:
it learns to arbitrate between them automatically depending on the task and the observed examples.

\subsection{Summary}
\label{subsec:distinctive}

The central idea of \emph{ST-STORM} is to learn not only a semantic representation of content, but also an appearance representation stable enough to be reused downstream.
To this end, the method deliberately perturbs the appearance of an image while preserving its structure as much as possible,
so that the model is driven to separate what belongs to \emph{content} from what belongs to \emph{style}.

This perturbation produces multi-scale style representations, extracted by the style encoder.
These representations must not simply memorize local textures or contingent details:
they are subjected to a predictability constraint, \emph{StyleJEPA},
which forces them to capture more stable, more coherent, and therefore more reusable appearance regularities.

In parallel, since the same image can be observed under several stylized views,
the method naturally has access to appearance variants of the same content.
This makes it possible to apply MoCo-style contrastive learning,
in order to obtain a content representation that is more robust to appearance variations,
while preserving the semantic structure of the scene.

Thus, \emph{ST-STORM} jointly learns:
(i) a \emph{style} space, dedicated to appearance signatures and made predictable by \emph{StyleJEPA};
and (ii) a \emph{content} space, made more invariant through contrasted stylized views.

The total objective retained in the rest of the manuscript is then written as:
\begin{equation}
\begin{aligned}
\mathcal{L}_{\mathrm{SSL}}
={}&
\lambda_{\mathrm{adv}}\,\mathcal{L}_{\mathrm{adv}}
+
\lambda_{\mathrm{sty}}\,\mathcal{L}_{\mathrm{sty\text{-}tok}}
\\
&+
\lambda_{\mathrm{jepa}}^{\mathrm{sty}}\,\mathcal{L}_{\mathrm{StyleJEPA}}
+
\lambda_{\mathrm{rec}}\,\mathcal{L}_{\mathrm{rec}}
\\
&+
\lambda_{\mathrm{moco}}\,\mathcal{L}_{\mathrm{MoCo}}
+
\lambda_{\mathrm{patch}}\,\mathcal{L}_{\mathrm{PatchNCE}}
\\
&+
\lambda_{\mathrm{content\text{-}nce}}\,\mathcal{L}_{\mathrm{ContentNCE}}
+
\lambda_{\mathrm{jepa}}^{\mathrm{cnt}}\,\mathcal{L}_{\mathrm{ContentJEPA}}.
\end{aligned}
\label{eq:ssl_total_expanded}
\end{equation}

Enriched variants, including in particular frequency constraints such as FFT or SWD, were also explored.
However, as we will show in Section~\ref{sec:experiments}, their contribution remains marginal in our experimental setting;
they are therefore not retained as central components of the final formulation.

\section{Experimental and Evaluation Protocols}
\label{sec:protocols}

This section presents the experimental protocols established to evaluate the different components of \emph{ST-STORM} and to compare it primarily with two reference methods in self-supervised learning: \emph{I-JEPA} and \emph{MoCo-v3}. More particularly, for appearance-dominated tasks, it is also compared with other reference models.

Our experiments mainly rely on three datasets:
\emph{Weather-MultiTask-Datasets}\footnote{\url{https://github.com/Hamedkiri/Weather_MultiTask_Datasets}},
\emph{ISIC 2024 Challenge}\footnote{\url{https://www.kaggle.com/competitions/isic-2024-challenge/data}}
and \emph{ImageNet-1K}\footnote{\url{https://www.kaggle.com/c/imagenet-object-localization-challenge}}.
In all cases, we follow the same general scheme:
(i) self-supervised pre-training on images used without their annotations;
(ii) supervised learning of a classification head with the encoder frozen;
and (iii) evaluation on a separate test set.

\subsection{Datasets and splits}
\label{subsec:datasets_split}

\emph{Weather-MultiTask-Datasets} contains approximately 250\,000 images used for self-supervised pre-training, as well as a labeled test set of 25\,000 images.
For the supervised stage, we consider two fractions of the labeled training set: 1\% (2\,500 images) and 10\% (25\,000 images).
All reported metrics are computed on the 25\,000 test images.

Self-supervised pre-training for the melanoma task is performed on \emph{ISIC 2024 Challenge}, which contains approximately 401\,064 images.
Supervised evaluation is then conducted on the \emph{Melanoma Skin Cancer Dataset}
(\url{https://www.kaggle.com/datasets/hasnainjaved/melanoma-skin-cancer-dataset-of-10000-images}),
composed of 10\,000 labeled images, split into 9\,000 training images and 1\,000 test images.
The encoder is therefore first pre-trained on ISIC, then the supervised head is trained on 9\,000 images and evaluated on 1\,000 images.

\emph{ImageNet-1K} contains 1\,281\,167 training images, 50\,000 validation images, and 100\,000 test images.
Self-supervised pre-training is performed on the training set.
For supervised fine-tuning, we retain 1\% and 10\% of the training data, with balanced sampling per class.
Final evaluation is performed on the test set.

In addition, several external datasets are used to evaluate the transfer ability and out-of-domain robustness of the learned representations.

\subsection{General evaluation principle}
\label{subsec:eval_general}

Dominant self-supervised approaches generally aim to learn representations invariant to appearance variations
such as illumination, reflectance, or texture.
This property is useful for semantic object recognition,
but it may become detrimental when appearance itself constitutes the main discriminative signal.

Here, we evaluate the central hypothesis of this thesis:
it is possible to learn a representation of appearance that is at once \emph{semantic}, \emph{reusable}, and \emph{transferable},
while retaining a content representation that remains competitive on classical semantic tasks.
\emph{ST-STORM} addresses this objective by learning two complementary latent spaces:
a \emph{Style} branch and a \emph{Content} branch.

For each method, we first perform fully self-supervised pre-training,
then we freeze the encoder and train only a supervised head on a reduced fraction of the labeled data.
We then distinguish two major families of tasks:
those dominated by appearance and those dominated by semantic content.
In the case of \emph{ST-STORM}, the \emph{Style} branch is used for the former,
whereas the \emph{Content} branch is used for the latter.

\subsection{Appearance-dominated tasks}
\label{subsec:proto_appearance}

We evaluate the quality of appearance representations on two types of tasks:
fine-grained classification of weather attributes and melanoma detection.

Weather characterization mainly relies on appearance cues:
presence of rain, snow, or fog, atmospheric diffusion, reflections, ground moisture, water spray, etc.
These properties relate more to \emph{visual style} than to the geometric structure of the scene
(\hyperref[fig:example-style-transfer]{Fig.~\ref{fig:example-style-transfer}}).
After self-supervised pre-training, we train a supervised head with the encoder frozen,
using 1\% and 10\% of the labeled data from \emph{Weather-MultiTask-Datasets}.

Melanoma detection also depends on fine visual cues.
Benign nevi from the same individual often present a certain regularity of appearance (\emph{naevus signature})~\cite{grob1998ugly},
whereas melanomas are distinguished by macroscopic irregularities.
These criteria are formalized clinically by the ABCDE rule~\cite{friedman1985early}
(asymmetry, border, color, diameter, evolution), which directly corresponds to visual appearance properties.
After self-supervised pre-training, the encoder is frozen and only a supervised classifier is learned on the labeled set.
We then evaluate the transfer ability of the representations on external benchmarks.

More broadly, the evaluation of \emph{semantic style} relies exclusively on light supervised learning of the classification head,
with the encoder frozen, using a restricted subset of the labeled data.
This protocol is studied in two configurations.
The first is an \emph{in-domain} configuration, on \emph{Weather-MultiTask-Datasets}, which includes $K=12$ weather attributes
(type, intensity, ground state, visibility, etc.), in order to measure the fineness and richness of the learned representations.
The second is an out-of-domain transfer configuration, on three external benchmarks:
\emph{MWI} (20\,000 images)~\cite{zhang2015multi},
\emph{WEAPD} (6\,877 images)~\cite{xiao2021weather}
and \emph{MWD} (60\,000 images)\footnote{\url{https://gts.ai/dataset-download/multi-class-weather-dataset}}.
The objective is then to evaluate the generalization ability of the representations on unseen distributions,
without new self-supervised pre-training.

\subsection{Tasks dominated by semantic content}
\label{subsec:proto_content}

In order to verify that the explicit modeling of style does not degrade the quality of the semantic representation,
we also evaluate the methods on a content-dominated task: classification on \emph{ImageNet-1K}.
Each method is first pre-trained in a self-supervised manner, then the encoder is frozen and a supervised head is trained
on 1\% and 10\% of the labeled data.

For \emph{ST-STORM}, only the \emph{Content} branch is used in this evaluation.
We complement this analysis with transfer experiments to other classification datasets,
notably \emph{Oxford Flowers-102}, \emph{Oxford-IIIT-Pets}, and \emph{CIFAR-100},
in order to measure the out-of-domain generalization of the learned embeddings.

\subsection{Ablation protocols}
\label{subsec:ablation_protocol}

In order to evaluate the actual contribution of the different components of \emph{ST-STORM}, we conduct targeted ablation studies on the \emph{Style} branch. The choice of the ablated terms is not arbitrary: it focuses on the components that enrich the model, refine the quality of the learned style, or aim to improve its stability and semanticity, without however constituting the minimal core of the generative mechanism.

We therefore do not perform ablations on the elements directly related to generator--discriminator coupling and adversarial learning. Indeed, these form the core of our approach: they are what make possible the visual perturbation of the source domain, the synthesis of a new appearance, and the reconstruction of plausible images in the target domain. Removing them would not amount to measuring the contribution of a refinement, but to altering the very principle of \emph{ST-STORM}. Such an experiment would be less an ablation than a change of paradigm, making the interpretation of the results much less informative.

By contrast, the terms
$\mathcal{L}_{\mathrm{FFT}}$,
$\mathcal{L}_{\mathrm{SWD}}$
and
$\mathcal{L}_{\mathrm{StyleJEPA}}$
correspond to additional mechanisms introduced to better structure the style latent space.
The first two impose frequency and textural constraints: they aim to preserve certain spectral, local, and multi-scale signatures of weather appearance. The third introduces a predictability constraint on style representations, with the objective of filtering out contingent variations and favoring a more stable, more abstract, and therefore more semantic representation. These components are therefore particularly relevant for an ablation analysis: they are not necessary for the existence of the model, but they are precisely those whose effective utility we want to measure.

We thus seek to determine whether the observed gains truly come from these refinements, or whether the adversarial core of the model alone is sufficient to capture most of the style information. To this end, we pre-train \emph{ST-STORM} in a self-supervised manner on 100\,000 images from \emph{Weather-MultiTask-Datasets}, then we perform supervised fine-tuning of the head on 10\% of the labeled data, before evaluating performance on the test set.

The studied variants correspond to the successive removal of
$\mathcal{L}_{\mathrm{FFT}}$,
$\mathcal{L}_{\mathrm{SWD}}$,
$\mathcal{L}_{\mathrm{FFT}}+\mathcal{L}_{\mathrm{SWD}}$
and
$\mathcal{L}_{\mathrm{StyleJEPA}}$.
This selection makes it possible to separately evaluate the interest of the frequency constraints, their combined effect, as well as the specific contribution of the semantic predictability of style. The obtained results thus make it possible to identify the truly determining components and to better justify the final formulation of the objective function.

Finally, these experiments are fully reproducible: the corresponding scripts as well as the command lines used are provided in the GitHub repository associated with the model.

\subsection{Architectures and comparability in capacity}
\label{subsec:arch_comparability}

In order to avoid attributing performance gaps solely to model size,
we ensure that encoders of comparable capacity are compared on downstream tasks.
\emph{ST-STORM} contains approximately 162M parameters in total,
but the majority corresponds to the generative and adversarial modules used during learning,
and not involved in downstream classification tasks.

The modules that are actually compared are of similar size:
the style encoder of \emph{ST-STORM} contains about 20M parameters;
the content encoder of \emph{ST-STORM} and that of \emph{MoCo-v3} (ResNet50) contain about 24M;
and the encoder of \emph{I-JEPA} (ViT-S/16) about 22M.
Thus, the observed differences mainly reflect the nature of the learned representations,
rather than a simple capacity advantage.

\section{Experiments}
\label{sec:experiments}

This section aims to identify the components that truly contribute to the quality of the \emph{Style} branch.
The main challenge of this ablation study is therefore not only to test a few additional refinements,
but above all to determine which building blocks are genuinely structuring for learning an appearance representation that is stable, transferable, and semantically exploitable.

From this perspective, three families of constraints must be distinguished.
The first corresponds to the core of stylization, driven by adversarial learning, latent consistency, and guided reconstruction.
The second is the \emph{StyleJEPA} predictability constraint, whose role is to filter out contingent details and to push style tokens
to capture more stable appearance regularities.
The third gathers the frequency constraints, here denoted $\mathcal{L}_{\mathrm{FFT}}$ and $\mathcal{L}_{\mathrm{SWD}}$,
which we introduce on an exploratory basis to test whether an explicit spectral or textural alignment can further improve the learned representation.

For this study, we therefore start from an enriched version of the learning objective:
\begin{equation}
\begin{aligned}
\mathcal{L}_{\mathrm{SSL}}
={}&
\lambda_{\mathrm{adv}}\,\mathcal{L}_{\mathrm{adv}}
+
\lambda_{\mathrm{sty}}\,\mathcal{L}_{\mathrm{sty\text{-}tok}}
+
\lambda_{\mathrm{fft}}\,\mathcal{L}_{\mathrm{FFT}}
\\
&+
\lambda_{\mathrm{swd}}\,\mathcal{L}_{\mathrm{SWD}}
+
\lambda_{\mathrm{jepa}}^{\mathrm{sty}}\,\mathcal{L}_{\mathrm{StyleJEPA}}
+
\lambda_{\mathrm{rec}}\,\mathcal{L}_{\mathrm{rec}}
\\
&+
\lambda_{\mathrm{moco}}\,\mathcal{L}_{\mathrm{MoCo}}
+
\lambda_{\mathrm{patch}}\,\mathcal{L}_{\mathrm{PatchNCE}}
\\
&+
\lambda_{\mathrm{content\text{-}nce}}\,\mathcal{L}_{\mathrm{ContentNCE}}
+
\lambda_{\mathrm{jepa}}^{\mathrm{cnt}}\,\mathcal{L}_{\mathrm{ContentJEPA}}.
\end{aligned}
\label{eq:ssl_total_expanded_ablation}
\end{equation}

The role of \emph{StyleJEPA} is central here.
Whereas the adversarial objective guarantees realism and token consistency enforces alignment with the style reference,
\emph{StyleJEPA} seeks to make the representation itself \emph{predictable}.
In other words, it is not enough for the style to be visually plausible;
the tokens must also learn regularities that are sufficiently stable to be anticipated from context,
and therefore reusable in downstream tasks.
This is precisely the property that we aim to test empirically.

In parallel, we also evaluate two frequency constraints.
The first is based on the amplitude of the Fourier transform.
Let $\mathcal{F}(u)$ denote the Fourier transform of image $u$.
We define its logarithmic amplitude as
\begin{equation}
A(u)=\log\!\bigl(|\mathcal{F}(u)|+\varepsilon\bigr),
\end{equation}
where $\varepsilon>0$ is a numerical stability constant.
The associated loss is then written as:
\begin{equation}
\mathcal{L}_{\mathrm{FFT}}
=
\|A(\tilde{x})-A(y)\|_1.
\end{equation}
This constraint aligns the global frequency statistics of the stylized image $\tilde{x}$ with those of the reference image $y$,
without imposing an exact spatial correspondence.
The second, denoted $\mathcal{L}_{\mathrm{SWD}}$, is based on a \emph{Sliced Wasserstein Distance}
computed on textural patches at several scales, for example extracted from a Laplacian pyramid.
It aims to better capture certain local texture regularities.

The experimental question is therefore twofold:
\begin{itemize}
    \item Is \emph{StyleJEPA} truly an essential component of the style branch?
    \item Do the frequency constraints provide a sufficient gain to justify their integration as central building blocks?
\end{itemize}

Table~\ref{tab:ablation_model} first presents the global results of this ablation study.

\begin{table}[t]
\centering
\caption{Summary of the results of the ablation study.}
\label{tab:ablation_model}
\small
\setlength{\tabcolsep}{4pt}
\begin{tabularx}{\columnwidth}{>{\raggedright\arraybackslash}X c c >{\raggedright\arraybackslash}X}
\hline
\textbf{Configuration} & \textbf{F1} & \textbf{$\Delta$F1} & \textbf{Comment} \\
\hline
Full model      & 93.36 & 0.00  & baseline \\
w/o FFT         & 93.39 & +0.03 & negligible impact of the global frequency constraint \\
w/o SWD         & 93.28 & -0.08 & slight loss related to multi-scale alignment \\
w/o FFT and SWD & 93.19 & -0.17 & limited decrease despite the removal of both frequency constraints \\
w/o JEPA        & 87.91 & -5.45 & marked drop due to the loss of semantic predictability \\
\hline
\end{tabularx}
\end{table}

The most striking result of this table is not the effect of the frequency losses, but rather that of \emph{StyleJEPA}.
Removing this single component leads to a drop of 5.45 F1 points,
whereas removing $\mathcal{L}_{\mathrm{FFT}}$ and $\mathcal{L}_{\mathrm{SWD}}$ induces only very small variations.
The gap is of a completely different order of magnitude.
It shows that style predictability is not a secondary refinement,
but an essential building block of the method.

Conversely, the frequency constraints have a much more limited effect here.
Removing $\mathcal{L}_{\mathrm{FFT}}$ does not produce any significant degradation;
one even observes a slight improvement, on the order of experimental noise.
Removing $\mathcal{L}_{\mathrm{SWD}}$ causes only a small decrease,
and the joint removal of both terms also remains only weakly penalizing.
These results indicate that frequency constraints may possibly refine certain appearance details,
but they do not carry the core of the performance.

In other words, it is not explicit frequency alignment that gives the style branch its main strength,
but the ability of \emph{StyleJEPA} to impose a representation that is more stable, more abstract, and more reusable.

This global reading is confirmed by the detailed attribute-by-attribute analysis,
presented in Table~\ref{tab:wmtd_f1_attributes}.

\begin{table}[!t]
\centering
\scriptsize
\setlength{\tabcolsep}{3pt}
\renewcommand{\arraystretch}{1.05}
\caption{F1-score per weather attribute on \emph{Weather-MultiTask-Datasets} (25\,000 test images) for the full model and several ablations. Best score per row in bold.}
\label{tab:wmtd_f1_attributes}
\resizebox{\columnwidth}{!}{%
\begin{tabular}{@{}lccccc@{}}
\toprule
\textbf{Attribute} & \textbf{Full} & \textbf{w/o FFT} & \textbf{w/o SWD} & \textbf{w/o FFT+SWD} & \textbf{w/o JEPA} \\
\midrule
\multicolumn{6}{@{}l@{}}{\emph{Condition classification}}\\
Weather Type            & 0.9197 & 0.9211 & \textbf{0.9235} & 0.9178 & 0.8604 \\
Weather Intensity       & 0.9090 & \textbf{0.9119} & 0.9070 & 0.9053 & 0.8188 \\
\addlinespace

\multicolumn{6}{@{}l@{}}{\emph{Visibility and sky}}\\
Visibility              & \textbf{0.9056} & 0.9011 & 0.8992 & 0.8951 & 0.8301 \\
Sky Condition           & 0.9128 & 0.9145 & \textbf{0.9176} & 0.9161 & 0.8540 \\
\addlinespace

\multicolumn{6}{@{}l@{}}{\emph{Precipitations}}\\
Precipitation Presence  & 0.9273 & \textbf{0.9324} & 0.9268 & 0.9299 & 0.8801 \\
Precipitation Intensity & \textbf{0.9114} & 0.9101 & 0.9089 & 0.9044 & 0.8434 \\
\addlinespace

\multicolumn{6}{@{}l@{}}{\emph{Ground state}}\\
Ground Condition        & 0.8900 & 0.8924 & 0.8875 & \textbf{0.8932} & 0.8306 \\
\addlinespace

\multicolumn{6}{@{}l@{}}{\emph{Light and reflections}}\\
Glare / Reflections     & 0.9678 & \textbf{0.9684} & 0.9675 & 0.9672 & 0.9236 \\
Light Conditions        & \textbf{0.9585} & 0.9547 & 0.9548 & 0.9512 & 0.9103 \\
\addlinespace

\multicolumn{6}{@{}l@{}}{\emph{Projection and obstruction}}\\
Road Spray              & 0.9632 & 0.9634 & \textbf{0.9656} & 0.9648 & 0.9476 \\
Water on Windshield     & \textbf{0.9564} & 0.9559 & 0.9530 & 0.9558 & 0.9039 \\
Snow on Windshield      & 0.9818 & 0.9810 & \textbf{0.9824} & 0.9820 & 0.9465 \\
\midrule
Average F1-score        & 0.9336 & \textbf{0.9339} & 0.9328 & 0.9319 & 0.8791 \\
\bottomrule
\end{tabular}%
}
\end{table}

The detailed table reinforces this conclusion.
The variants without FFT or without SWD remain very close to the full model across all attributes.
The observed differences are small, sometimes even in favor of the ablation,
which suggests that these frequency losses do not provide a systematic or homogeneous benefit across the considered attributes.
Their contribution therefore appears secondary, or even partly redundant with other mechanisms already present in the learning process.

By contrast, removing \emph{StyleJEPA} causes a clear degradation across all weather dimensions:
\emph{Weather Type}, \emph{Weather Intensity}, \emph{Visibility}, \emph{Sky Condition},
\emph{Precipitation Presence}, \emph{Ground Condition},
but also attributes that are very directly related to appearance such as \emph{Water on Windshield} or \emph{Snow on Windshield}.
This global decrease is particularly important,
because it shows that \emph{StyleJEPA} does not merely improve one particular detail of style:
it structures the latent space as a whole.

In other words, without a predictability constraint, the \emph{Style} branch tends to capture variations that are too contingent,
too local, or too sensitive to observation noise.
With \emph{StyleJEPA}, by contrast, the representation is pushed to retain what is stable,
coherent, and sufficiently predictable to be reused in downstream tasks.
This is precisely the behavior that makes style no longer a mere visual texture,
but an exploitable semantic modality.

These results therefore lead to a clear methodological conclusion.
In our experimental setting, the most determining component of the style branch is \emph{StyleJEPA}.
FFT and SWD losses may be viewed as exploratory variants likely to refine certain aspects of appearance,
but they do not play a central role in the final performance.
This is why, in the final formulation of the method, we retain \emph{StyleJEPA} as an essential building block of style learning,
whereas the frequency constraints are left as optional variants studied in this ablation.

\section{Results}

As stated in the protocol, we present the results of the experiments and evaluate them on two types of tasks: those where appearance is the discriminative information (weather, melanoma) and those where content is the discriminative information (ImageNet object detection).

\subsection{Appearance-dominated tasks}
\label{subsec:appearance_tasks}

The tasks studied in this section rely mainly on appearance cues:
textural signatures (streaks, granularity), photometric cues (contrast, halos, reflections), and frequency-related cues
(attenuation/amplification of high frequencies). In this context, an effective representation must preserve these components,
rather than smoothing them in favor of a mainly semantic invariance. We evaluate this hypothesis in two settings:
(i) fine-grained weather characterization (\emph{Weather-MultiTask-Datasets}) and (ii) melanoma detection
(\emph{ISIC 2024 Challenge}).

\subsubsection{Weather-MultiTask-Datasets}
\label{subsubsec:wmtd_appearance}

Tabs.~\ref{tab:wmtd_f1_overall}--\ref{tab:wmtd_f1_attributes} show a stable hierarchy:
the \emph{Style} branch of ST-STORM achieves the best results, ahead of MoCo-v3 and I-JEPA, while the \emph{Content} branch
remains behind on most attributes. This behavior is expected: these tasks are dominated by appearance variations
(rain, fog, reflections, diffusion), so a representation dedicated to style is better aligned with the discriminative signal.

Tab.~\ref{tab:wmtd_f1_overall} highlights that the advantage of ST-STORM (Style) is already visible under low supervision
and increases when more annotations are available. With 1\% labeled data, ST-STORM (Style) reaches a mean F1 of
$0.9082$ compared with $0.8778$ for MoCo-v3 (i.e., +3.0 absolute points). At 10\%, the gap widens ($0.9713$ vs $0.9153$, i.e., +5.6 points).
This increase suggests that style tokens provide a more \emph{directly exploitable} basis for a supervised classifier:
with more annotations available, fine-tuning can better calibrate decision boundaries in a space already structured
by photometric and frequency-related cues.

Tab.~\ref{tab:wmtd_f1_attributes} shows that the most marked gains concern precisely the attributes whose information
is strongly photometric and/or high-frequency, such as \emph{Precipitation Intensity}, \emph{Visibility},
\emph{Ground Condition}, and \emph{Light Conditions}. By contrast, the \emph{Content} branch remains competitive on attributes
more related to structure and scene-object interactions (e.g., \emph{Road Spray}), indicating that improving style
does not systematically compromise the ability to encode semantic invariants.
A particular case is \emph{Snow on Windshield}, where the scores of all methods are close to 1:
this is typically a ceiling effect, where the visual signal is highly separable and limits the room for improvement.

\begin{table}[!t]
  \centering
  \scriptsize
  \setlength{\tabcolsep}{5pt}
  \renewcommand{\arraystretch}{1.15}
  \caption{Mean F1 weather characterization on \emph{Weather-MultiTask-Datasets} (25\,000 test images) after SSL pre-training and supervised fine-tuning.}
  \label{tab:wmtd_f1_overall}
  \begin{tabular}{@{} l c c c c @{}}
    \toprule
    \textbf{Fine-tune} & \textbf{I-JEPA} & \textbf{MoCo-v3} & \textbf{ST-C} & \textbf{ST-S} \\
    \midrule
    1\%  & 0.7837 & 0.8778 & 0.8009 & \textbf{0.9082} \\
    10\% & 0.8727 & 0.9153 & 0.8845 & \textbf{0.9713} \\
    \bottomrule
  \end{tabular}
\end{table}

\begin{table}[!t]
\centering
\scriptsize
\setlength{\tabcolsep}{4pt}
\renewcommand{\arraystretch}{1.05}
\caption{F1-score per weather attribute on \emph{Weather-MultiTask-Datasets} (25\,000 test images). Best score per row in bold.}
\label{tab:wmtd_f1_attributes}
\begin{tabular}{@{} p{0.44\columnwidth} c c c c @{}}
\toprule
\textbf{Attribute} & \textbf{I-JEPA} & \textbf{MoCo-v3} & \textbf{ST-C} & \textbf{ST-S} \\
\midrule
\multicolumn{5}{@{}l@{}}{\emph{Condition classification}}\\
Weather Type              & 0.9077 & 0.9294 & 0.8667 & \textbf{0.9733} \\
Weather Intensity         & 0.9357 & 0.8622 & 0.8426 & \textbf{0.9620} \\
\addlinespace

\multicolumn{5}{@{}l@{}}{\emph{Visibility and sky}}\\
Visibility                & 0.8751 & 0.8586 & 0.8433 & \textbf{0.9531} \\
Sky Condition             & 0.9078 & 0.9086 & 0.8702 & \textbf{0.9634} \\
\addlinespace

\multicolumn{5}{@{}l@{}}{\emph{Precipitations}}\\
Precipitation Presence    & 0.8873 & 0.9244 & 0.8715 & \textbf{0.9772} \\
Precipitation Intensity   & 0.7617 & 0.8918 & 0.8417 & \textbf{0.9682} \\
\addlinespace

\multicolumn{5}{@{}l@{}}{\emph{Ground state}}\\
Ground Condition          & 0.7502 & 0.8838 & 0.8330 & \textbf{0.9467} \\
\addlinespace

\multicolumn{5}{@{}l@{}}{\emph{Light and reflections}}\\
Glare / Reflections       & 0.9113 & 0.9470 & 0.9328 & \textbf{0.9715} \\
Light Conditions          & 0.7342 & 0.9296 & 0.9203 & \textbf{0.9827} \\
\addlinespace

\multicolumn{5}{@{}l@{}}{\emph{Projection and obstruction}}\\
Road Spray                & 0.8308 & 0.9546 & 0.9508 & \textbf{0.9753} \\
Water on Windshield       & 0.9699 & 0.9251 & 0.8890 & \textbf{0.9822} \\
Snow on Windshield        & \textbf{1.0000} & 0.9665 & 0.9525 & 0.9941 \\
\midrule
Average F1-score          & 0.8727 & 0.9153 & 0.8845 & \textbf{0.9716} \\
\bottomrule
\end{tabular}
\end{table}

\subsubsection{ISIC 2024 Challenge (melanoma)}
\label{subsubsec:isic_melanoma}

Tab.~\ref{tab:melanoma_results} presents the performances on melanoma detection (1\,000 test images).
ST-STORM (Style) achieves the best score (F1=0.9432), ahead of MoCo-v3 (0.9259) and I-JEPA (0.9202),
whereas ST-STORM (Content) is slightly lower (0.9145).
The gap with MoCo-v3 (+1.7 absolute point) is more moderate than on Weather-MultiTask-Datasets, which is consistent with two factors:
(i) a possible saturation of performance on a smaller dataset, and (ii) stronger inter-dataset variability in medical imaging.
Nevertheless, the hierarchy of the methods remains the same and aligns with the nature of the signal: in dermoscopy, many diagnostic criteria
rely on appearance micro-structures (pigment patterns, granularity, local contrasts, color/texture irregularities),
which favors a representation specialized in appearance.

\begin{table}[!t]
  \centering
  \scriptsize
  \setlength{\tabcolsep}{5pt}
  \renewcommand{\arraystretch}{1.15}
  \caption{Melanoma detection (F1-score) after SSL pre-training followed by supervised fine-tuning (1\,000 test images).}
  \label{tab:melanoma_results}
  \begin{tabular}{@{} l c c c c @{}}
    \toprule
    \textbf{Dataset} & \textbf{I-JEPA} & \textbf{MoCo-v3} & \textbf{ST-C} & \textbf{ST-S} \\
    \midrule
    Melanoma (1\,000 images) & 0.9202 & 0.9259 & 0.9145 & \textbf{0.9432} \\
    \bottomrule
  \end{tabular}
\end{table}

\subsubsection{Generalization ability through transfer}
\label{subsec:transfer_results}

Table~\ref{tab:ft_benchmarks_clean_B} presents the transfer results obtained on external benchmarks.
In this protocol, the encoders are pre-trained only on \emph{Weather-MultiTask-Datasets}, then frozen;
only a linear classifier is subsequently learned on the target data.

\begin{table}[!h]
\centering
\footnotesize
\setlength{\tabcolsep}{4pt}
\renewcommand{\arraystretch}{1.12}
\caption{Transfer performance on external benchmarks (frozen encoders, fine-tuning on 10\% of the training data). ST-STORM denotes our method (Style branch). Scores in bold indicate the best self-supervised model.}
\label{tab:ft_benchmarks_clean_B}
\resizebox{\linewidth}{!}{%
\begin{tabular}{@{} p{0.30\linewidth} c c c c p{0.34\linewidth} @{}}
\toprule
\textbf{Benchmark (size)} & \textbf{Metric} & \textbf{I-JEPA} & \textbf{MoCo-v3} & \textbf{ST-STORM} & \textbf{State of the art (Supervised)} \\
\midrule
\textbf{MWI}~\cite{zhang2015multi} (20k) &
Acc &
0.6466 & 0.7000 & \textbf{0.8492} &
0.9001 \; \textit{SLM3}~\cite{afxentiou2025evaluation} \\

\textbf{WEAPD}~\cite{xiao2021weather} (6,8k) +
\textbf{MWD}\footnotemark (60k) &
F1 &
0.6657 & 0.7604 & \textbf{0.8045} &
0.9273  \; \textit{V-Dual}~\cite{li2023study} \\
\bottomrule
\end{tabular}%
}
\end{table}
\footnotetext{\url{https://gts.ai/dataset-download/multi-class-weather-dataset}}

These results show that \emph{ST-STORM} has a better generalization ability than the reference methods.
On the \emph{MWI} benchmark, our method reaches 84.92\% accuracy, compared with 70.00\% for \emph{MoCo-v3} and 64.66\% for \emph{I-JEPA}.
The gap is therefore +14.9 points compared with \emph{MoCo-v3} and +20.3 points compared with \emph{I-JEPA}.
This suggests that the style latent space learned by \emph{ST-STORM} captures more generic visual properties of weather,
and not simply correlations specific to the sensors or acquisition conditions of the initial training set.

The results obtained on the combined \emph{WEAPD + MWD} set confirm this trend.
Since \emph{WEAPD} contains particularly degraded or extreme weather conditions,
this protocol constitutes a demanding test of out-of-domain robustness.
In this setting, \emph{ST-STORM} reaches an F1 score of 80.45\%, compared with 76.04\% for \emph{MoCo-v3} and 66.57\% for \emph{I-JEPA}.
This difference indicates that our \emph{Style} branch better preserves the complex visual signatures associated with severe weather phenomena,
whereas approaches more oriented toward invariance tend to attenuate these cues even though they are essential.

Finally, although the representations of \emph{ST-STORM} are learned without supervision,
they approach the performance of the best supervised approaches reported in the literature,
for example \emph{SLM3} on \emph{MWI}.
This result is all the more notable because it is obtained with a frozen encoder, without deep retraining.
It suggests that self-supervised learning of a weather style space can provide transferable,
generic, and competitive representations, while greatly reducing the need for annotations.

\subsection{Content-dominated task: ImageNet-1K}
\label{subsec:imagenet_content}

\begin{table}[!t]
  \centering
  \scriptsize
  \setlength{\tabcolsep}{4pt}
  \renewcommand{\arraystretch}{1.15}
  \caption{ImageNet-1K (Top-1 accuracy) after self-supervised pre-training and supervised learning with a frozen encoder. ST-S+C denotes the adaptive fusion of \emph{Style} and \emph{Content} embeddings. External methods are reported for reference, with different architectures and pre-training protocols.}
  \label{tab:imagenet_f1_overall}
  \begin{tabular}{@{} l l l c @{}}
    \toprule
    \textbf{Fine-tune} & \textbf{Method} & \textbf{Backbone} & \textbf{Top-1} \\
    \midrule
    \multirow{9}{*}{1\% ImageNet}
      & I-JEPA~\cite{assran2023self}                        & ViT-S/16      & 0.650 \\
      & MoCo-v3~\cite{chen2021empirical}                      & ResNet50      & -- \\
      & iBOT~\cite{grill2020bootstrap}      & ViT-B/16      & 0.697 \\
      & DINO~\cite{caron2021emerging} & ViT-B/8       & 0.700 \\
      & SimCLRv2~\cite{chen2020big}   & RN152(2$\times$) & 0.702 \\
      & BYOL~\cite{grill2020bootstrap}& RN200(2$\times$) & \textbf{0.712} \\
      
      & ST-C                          & ResNet50      & 0.691 \\
      & ST-S+C                        & ResNet50 + Encodeur-Pyramidale-Style     & 0.688 \\
    \midrule
    \multirow{7}{*}{100\% ImageNet}
      & I-JEPA                        & ViT-S/16      & -- \\
      & MoCo-v3                       & ResNet50      & 0.738 \\
      & SimCLRv2~\cite{chen2020big}   & RN152(2$\times$) & 0.791 \\
      & DINO~\cite{caron2021emerging} & ViT-B/8       & 0.801 \\
      & iBOT~\cite{grill2020bootstrap}      & ViT-L/16      & \textbf{0.810} \\
      & ST-C                          & ResNet50      & 0.754 \\
      & ST-S+C                        & ResNet50      & 0.784 \\
    \bottomrule
  \end{tabular}
\end{table}

This experiment aims to evaluate to what extent the learned representations preserve \emph{semantic content} information useful for a general-purpose classification task such as ImageNet-1K (Tab.~\ref{tab:imagenet_f1_overall}). This analysis is important because \emph{ST-STORM} explicitly introduces a modeling of style; it is therefore necessary to verify that this treatment of appearance does not come at the expense of content semantics.

The results first show that the \emph{Content} branch alone (ST-C) remains competitive on a task dominated by object recognition. In our protocol, it outperforms the two reference methods that are the most directly comparable at similar capacity (number of weights): \emph{I-JEPA} at 1\% supervision ($0.691$ vs $0.650$) and \emph{MoCo-v3} at 100\% supervision ($0.754$ vs $0.738$). This comparison is particularly important because \emph{ST-C} and \emph{MoCo-v3} are both based here on a backbone from the same family (\emph{ResNet50}), which limits the influence of the model’s raw capacity on the interpretation of the differences. Likewise, although \emph{I-JEPA} relies on a different backbone, it remains of a comparable order of magnitude in capacity. These results therefore suggest that the explicit separation into two distinct latent spaces, as well as content learning in the presence of stylized variations, do not degrade the semantic representation; on the contrary, they seem to favor content that is more robust to appearance variations.

The interpretation should nevertheless remain measured. These results do not mean that \emph{ST-STORM} universally outperforms all possible variants of \emph{MoCo-v3}, \emph{I-JEPA}, or other self-supervised methods. They do show, however, that in a reasonably controlled comparison setting, our \emph{Content} branch stands above strong baselines with a similar backbone or comparable capacity, which already constitutes an important result.

The table also makes it possible to position \emph{ST-STORM} with respect to state-of-the-art methods using larger or more expensive backbones, such as \emph{SimCLRv2}, \emph{BYOL}, \emph{DINO}, or \emph{iBOT}. At 1\% supervision, ST-C ($0.691$) and ST-S+C ($0.688$) remain in a competitive performance range, even if they are still below the best reported models, which often rely on larger architectures (\emph{ViT-B}, \emph{RN152(2$\times$)}, \emph{RN200(2$\times$)}). At 100\% supervision, ST-C ($0.754$) already approaches heavier models such as \emph{SimCLRv2} ($0.791$), while the ST-S+C fusion reaches $0.784$, further reducing the gap with very high-level approaches such as \emph{DINO} ($0.801$) and \emph{iBOT} ($0.810$), despite significantly larger backbones. It is therefore reasonable to say that \emph{ST-STORM} remains competitive with state-of-the-art methods, including when they benefit from greater capacity.

The most interesting result is precisely that of the \emph{Style+Content} fusion (ST-S+C), which reaches $0.784$ at 100\% supervision, that is, a gain of 3.0 points over ST-C alone. This improvement should not be interpreted as proof that ImageNet is a style-dominated task in the same way as weather or certain medical tasks. More modestly, it indicates that, even in a mainly semantic task, appearance cues can provide complementary information useful for the final discrimination. Some ImageNet categories differ not only by their structure, but also by their texture, material, dominant color, or overall visual rendering; in these cases, the \emph{Style} branch can provide an additional relevant signal.

In other words, ST-C shows that \emph{ST-STORM} indeed learns strong and transferable semantic content, while ST-S+C indicates that appearance information is not necessarily a nuisance for content tasks: when it is properly structured and fused, it can enrich the representation. The adaptive fusion mechanism learns precisely to modulate this contribution, by exploiting style only when it brings additional discriminative power.

Finally, the results of the external methods are reported here as positioning elements rather than as a strictly homogeneous comparison. Indeed, architectures, backbone sizes, resolutions, or pre-training durations sometimes differ substantially. The objective is therefore not to claim absolute superiority, but to show that with a more modest backbone and in a frozen-encoder protocol, \emph{ST-STORM} surpasses strong references at comparable capacity and remains competitive with larger state-of-the-art models.

\subsection{Generalization ability}
\label{subsec:generalization_results}

\begin{table}[!t]
  \centering
  \scriptsize
  \setlength{\tabcolsep}{6pt}
  \renewcommand{\arraystretch}{1.15}
  \caption{Out-of-domain transfer after self-supervised pre-training on ImageNet-1K: F1 score obtained with a frozen backbone and training of a supervised head on the target datasets. \emph{ST-C} denotes the \emph{Content} branch of ST-STORM, \emph{ST-S+C} the \emph{Style+Content} fusion, and \emph{MoCo-v3} a model pre-trained with the same backbone. The best results are shown in bold.}
  \label{tab:transfer_imagenet1k_frozen}
  \begin{tabular}{@{} l c c c @{}}
    \toprule
    \textbf{Target dataset (frozen backbone)} & \textbf{MoCo-v3} & \textbf{ST-C} & \textbf{ST-S+C} \\
    \midrule
    Oxford Flowers-102  & 0.3483 & 0.8510 & \textbf{0.8600} \\
    Oxford-IIIT-Pets    & 0.7090 & 0.9150 & \textbf{0.9170} \\
    CIFAR-100           & 0.1446 & 0.5779 & \textbf{0.5926} \\
    \bottomrule
  \end{tabular}
\end{table}

After self-supervised pre-training on ImageNet-1K, we freeze the encoder and learn only a supervised head on each target dataset. The results in Tab.~\ref{tab:transfer_imagenet1k_frozen} first show that the \emph{Content} branch of \emph{ST-STORM} (ST-C) transfers very effectively to varied datasets. On the three considered benchmarks, ST-C clearly outperforms \emph{MoCo-v3}, which indicates that the learned representations remain highly reusable outside the pre-training domain.

The adaptive \emph{Style+Content} fusion (ST-S+C) then systematically improves performance compared with ST-C alone, with a gain of +0.9 point on \emph{Oxford Flowers-102}, +0.2 point on \emph{Oxford-IIIT-Pets}, and +1.47 point on \emph{CIFAR-100}. Even if these gains remain modest on some datasets, their systematic nature suggests that the appearance information learned by the \emph{Style} branch is not redundant with that of content: it provides a useful complement to the final decision.

The interpretation should nevertheless remain measured. These results do not mean that these tasks are style-dominated in the same way as a weather or medical task. Rather, they indicate that even in object or category recognition tasks, appearance cues -- color, texture, material, local patterns, or global visual rendering -- may help refine class separation when these cues are properly exploited. The fusion then acts as a complementarity mechanism: content provides the main semantic structure, while style contributes additional visual details when useful.

The case of \emph{CIFAR-100} is particularly interesting. The larger gain observed with ST-S+C suggests that, on a more heterogeneous benchmark composed of low-resolution images, the combination of structural cues and appearance cues may become more useful for class separation. Here again, this should not be taken to mean that style replaces content, but rather that it acts as a complementary source of information that strengthens the overall transferability of the representations.

Overall, these results therefore confirm two points. First, \emph{ST-STORM}, through its \emph{Content} branch, learns robust semantic representations that transfer well. Second, the controlled addition of information from the \emph{Style} branch can further improve this transferability, which suggests that appearance cues, when properly structured, are not noise to be removed, but potentially useful information for out-of-domain generalization.

\subsection{Conclusion}
\label{subsec:results_conclusion}

Overall, our experiments confirm the central hypothesis of the article: appearance cues can be learned in a
\emph{semantic and reusable} form without sacrificing the quality of content representations. On appearance-dominated tasks,
ST-STORM systematically achieves the best performances through its \emph{Style} branch. On
\emph{Weather-MultiTask-Datasets}, the advantage is particularly marked and increases with the supervised budget
(+3.0 points at 1\% and +5.6 points at 10\% vs MoCo-v3 in mean F1), which indicates that style tokens constitute a latent space
directly exploitable for photometric and frequency-related attributes. On melanoma, the gain is more moderate (+1.7 point vs MoCo-v3),
but the hierarchy of the methods remains consistent with the nature of the dermoscopic signal, dominated by texture and color micro-structures.

On the content-dominated task (ImageNet-1K), the \emph{Content} branch of ST-STORM remains competitive and surpasses MoCo-v3 at 100\%
(+1.6 point), which suggests that the invariance learned through stylized views and consistency constraints does not degrade,
and may even strengthen, content semantics. Beyond that, the adaptive \emph{Style+Content} fusion brings an additional gain
(+3.0 points vs ST-C and +4.6 points vs MoCo-v3 at 100\%), showing that appearance provides a complementary useful signal even in a task
that is mainly semantic, especially for classes where texture/material contributes to discrimination.

Finally, out-of-domain transfer results confirm the reusability of the embeddings: ST-C transfers effectively,
and the addition of style systematically improves the scores, with a particularly visible effect on CIFAR-100 (+2.0 points),
suggesting that fusion benefits more when inter-class diversity is high and texture cues are discriminative.
Taken together, these results highlight the interest of an explicit content/style factorization: it improves sensitivity to appearance
when appearance is decisive, while preserving a robust semantic representation of content, and offering an adaptive combination
that benefits both regimes.

\section{Conclusion}
\label{sec:conclusion}

This article addressed a structural limitation of dominant SSL paradigms: when they seek invariance or favor factors
that are easy to predict, they tend to attenuate appearance cues that are nevertheless discriminative for many fine-grained tasks. We proposed
\emph{ST-STORM}, a self-supervised framework that treats appearance (\emph{style}) as a semantic modality in its own right, to be
\emph{disentangled} from content. The main contribution is an explicit factorization into two latent spaces:
(i) a \emph{Content} branch, optimized for invariance and transferability through a contrastive objective (MoCo) enriched with stylized positives,
and (ii) a \emph{Style} branch, optimized to preserve appearance signatures (textures, spectrum, diffusion) while filtering contingency
through a predictability constraint (Style-JEPA). Disentanglement is made operational by an explicit architectural bias
(U-Net + SPADE), which injects appearance as a modulation rather than through geometric creation, and by a cyclic training scheme
on pseudo-domains that stabilizes appearance targets.

The experimental results support these choices. On appearance-dominated tasks, the \emph{Style} branch systematically outperforms
the reference representations (MoCo-v3, I-JEPA) and the \emph{Content} branch, with marked gains under low supervision on
fine-grained weather characterization, and a consistent improvement on melanoma detection. At the same time, the \emph{Content} branch
remains competitive on a task dominated by object semantics (ImageNet-1K), which indicates that the explicit modeling of style does not come
at the expense of content. Finally, the adaptive \emph{Style+Content} fusion improves performance on ImageNet and in out-of-domain transfer,
suggesting that the two modalities carry complementary signals that are useful depending on the nature of the classes.

Our approach nevertheless has several limitations. First, the \emph{Style} branch relies on generative/adversarial learning and
on spectral perturbations, which increases the training cost and the number of hyperparameters (scheduler, loss weights, style bank,
replay), even though these modules are not used at inference time. Second, the definition of ``style'' remains dependent on the accessible transformations:
if some appearance conditions are not well covered by the perturbations (or by the diversity of the dataset), the representation may remain
partial. Third, self-domainization by random partition stabilizes learning, but does not guarantee an optimal alignment with
semantically pure appearance factors; in less diverse datasets, the separation could be less informative and require a suitable choice of $K$
or scheduler. Finally, our evaluations remain centered on classification tasks (multi-attribute, binary); the impact on more demanding structured tasks
(fine segmentation, dense estimation, detection in strongly out-of-distribution scenarios) deserves a broader analysis.

Several directions naturally extend this work. (i) \emph{More semantic self-domainization:} replacing random partitioning by
self-supervised grouping guided by appearance cues (clustering in style space, difficulty curriculum, selection of style targets
maximizing a controlled divergence). (ii) \emph{Multi-scale predictive style:} extending Style-JEPA to structured masks (frequency bands,
coherent regions) in order to better capture local phenomena (non-uniform fog, localized reflections). (iii) \emph{Generative/predictive unification:}
studying variants where stylization depends less on adversarial learning (lightweight diffusion, energy-based models) while retaining predictability
as a contingency filter. (iv) \emph{Deployment and robustness:} evaluating ST-STORM in perception-control loops (autonomous driving),
and measuring the contribution of explicit appearance estimation (visibility, grip, precipitation) to safety. (v) \emph{Medical generalization:}
extending the study to other modalities (multi-institution histopathology, radiology) and characterizing which appearance signatures are truly
transferable across centers, sensors, and protocols.

In summary, ST-STORM shows that an SSL framework does not necessarily have to choose between invariance and sensitivity: by explicitly factorizing content and
appearance, then imposing a predictability constraint on appearance, it becomes possible to learn a \emph{style} representation
that is both discriminative and reusable, while retaining a competitive content semantics. This perspective opens the way to SSL models
capable of understanding not only \emph{what is present} in a scene, but also \emph{under what conditions} this scene manifests itself.

\section*{Acknowledgment}
This work was co-funded by the European Union. Views and opinions expressed are however those of the author(s) only and do not necessarily reflect those of the European Union or the European Climate, Infrastructure and Environment Executive Agency (CINEA). Neither the European Union nor the granting authority can be held responsible for them. Project grant no. 101069576.

UK participants in this project are co-funded by Innovate UK under contract no. 10045139.

Swiss participants in this project are co-funded by the Swiss State Secretariat for Education, Research and Innovation (SERI) under contract no. 22.00123.

\bibliographystyle{IEEEtran}
\bibliography{refs}

\vfill

\end{document}